%% file: main.tex
\documentclass[letterpaper, 10 pt, conference]{ieeeconf}  % Comment this line out if you need a4paper

\IEEEoverridecommandlockouts                              % This command is only needed if 
\input{preamble}

\makeatletter
\let\NAT@parse\undefined
\makeatother

\usepackage{hyperref}
\hypersetup{citebordercolor=green}
\hypersetup{linkbordercolor=green}
\hypersetup{urlbordercolor=green}
\graphicspath{{figs/}}

\title{\LARGE \bf
Online Informative Path Planning for Active Information \\Gathering
of a 3D Surface
}
\author{Hai Zhu$^1$, Jen Jen Chung$^2$, Nicholas R.J. Lawrance$^2$, Roland Siegwart$^2$, and Javier Alonso-Mora$^1$%
\thanks{This work is supported in part by the U.S. Office of Naval Research Global (ONRG) NICOP-grant N62909-19-1-2027}% <-this % stops a space
\thanks{$^1$Department of Cognitive Robotics, Delft University of Technology {\tt\footnotesize $\{$h.zhu; j.alonsomora$\}$@tudelft.nl}}%
\thanks{$^2$Autonomous Systems Lab, ETH Zurich {\tt\footnotesize $\{$jenjen.chung; nicholas.lawrance$\}$@mavt.ethz.ch; rsiegwart@ethz.ch}}
}
\begin{document}

\maketitle              % typeset the title of the contribution

%%%%%%%%%%%%%%%%%%%%%%%%%%%%%%%%%%%%%%%%%%%%%%%%%%%%%%%%%%%%%%%%%%%%%%%%%%%
\begin{abstract}
    This paper presents an online informative path planning approach for active information gathering on three-dimensional surfaces using aerial robots. Most existing works on surface inspection focus on planning a path offline that can provide full coverage of the surface, which inherently assumes the surface information is uniformly distributed hence ignoring potential spatial correlations of the information field. In this paper, we utilize manifold Gaussian processes (mGPs) with geodesic kernel functions for mapping surface information fields and plan informative paths online in a receding horizon manner. Our approach actively plans information-gathering paths based on recent observations that respect dynamic constraints of the vehicle and a total flight time budget. We provide planning results for simulated temperature modeling for simple and complex 3D surface geometries (a cylinder and an aircraft model). We demonstrate that our informative planning method outperforms traditional approaches such as 3D coverage planning and random exploration, both in reconstruction error and information-theoretic metrics. We also show that by taking spatial correlations of the information field into planning using mGPs, the information gathering efficiency is significantly improved. 
\end{abstract}

\input{./secs/01_introduction.tex}

\input{./secs/02_related_work.tex}

\input{./secs/03_formulation.tex}

\input{./secs/04_method_1.tex}

\input{./secs/06_results.tex}

\input{./secs/07_conclusion.tex}

%%%%%%%%%%%%%%%%%%%%%%%%%%%%%%%%%%%%%%%%%%%%%%%%%%%%%%%%%%%%%%%%%%%%%%%%%%
% \section*{Acknowledgements}

% \newpage
%%%%%%%%%%%%%%%%%%%%%%%%%%%%%%%%%%%%%%%%%%%%%%%%%%%%%%%%%%%%%%%%%%%%%%%%%%
\bibliographystyle{IEEEtran}
\balance
\bibliography{ref}

\end{document}

%% file: preamble.tex
\usepackage{times}
\usepackage{amssymb}
\usepackage{amsmath}
\usepackage{amsthm}
\usepackage{graphics}
\usepackage{epsfig}
\usepackage{multirow}
\usepackage{nicefrac}
\usepackage{cases}          % For multi-line equations with cases and numbers
\usepackage{subcaption}
\usepackage{mathtools}
\usepackage{xcolor}
\usepackage{algorithm}
\usepackage{algorithmicx}
\usepackage[noend]{algpseudocode}
\usepackage{balance}
\usepackage{textcomp}
\usepackage{gensymb}

\newcommand{\tn}[1]{\textnormal{#1}}
\newcommand{\tb}[1]{\textbf{#1}}

\newcommand{\mat}[0]{\begin{bmatrix}}
\newcommand{\mate}[0]{\end{bmatrix}}
\newcommand{\vc}{\mathbf{c}}

\newcommand{\vp}{\mathbf{p}}

\newcommand{\vy}{\mathbf{y}}

\newcommand{\vI}{\mathbf{I}}

\newcommand{\cC}{\mathcal{C}}

\newcommand{\cG}{\mathcal{G}}

\newcommand{\cL}{\mathcal{L}}
\newcommand{\cM}{\mathcal{M}}

\newcommand{\cP}{\mathcal{P}}

\newcommand{\cT}{\mathcal{T}}

\newcommand{\cX}{\mathcal{X}}

\newcommand{\R}{\mathbb{R}}

\newcommand\abs[1]{\left|#1\right|}                 % Big abs, determinant

\DeclareMathOperator*{\argmax}{arg\,max}            % argmax
\newcommand{\ra}{\rightarrow}

\newcommand{\la}{\leftarrow}

%% file: secs/01_introduction.tex
%%%%%%%%%%%%%%%%%%%%%%%%%%%%%%%%%%%%%%%%%%%%%%%%%%%%%%%%%%%%%%%%%%%%%%%%%%%%%%%%
\section{Introduction}\label{sec:intro}
Deploying Unmanned Aerial Vehicles (UAVs) for structural inspection has become popular in recent years thanks for their low costs, high maneuverability, and the ability to operate in hard-to-access environments. Typical examples include those of wind turbine inspection \cite{Gu2020}, bridge surface reconstruction \cite{Chen2019} and damage evaluation \cite{Kim2015}, aircraft exterior screw detection \cite{Miranda2019}, and pipeline thermal diagnostics \cite{Workswell2017}. In these applications, a 3D model with known or partially known geometry of the structure to be inspected is given a priori, which is generally represented by a mesh. Then the UAVs are employed to gather information (e.g. cracks, temperatures) on the surface by carrying task-specific sensors, such as RGB cameras, thermal cameras, and laser scanners for different inspection missions. 

Driven by these practical applications, facilitating UAV path planning for autonomous inspection has drawn significant research efforts \cite{Bircher2016,bircher2017incremental}. While most existing works focus on planning a global path that can provide full coverage of the surface, two main disadvantages are observed: a) the method inherently assumes that the surface information is uniformly distributed, hence ignoring potential spatial correlations of the information field (e.g. temperature distribution); b) the path is planned offline, so it cannot actively integrate measurements obtained during execution into the planning to improve the performance.

To overcome the two issues, in this paper we propose an online informative path planning approach that can enhance data acquisition efficiency by taking account of spatial correlations of the information field, and planning the path online in a receding horizon manner. The approach builds upon previous works \cite{Hitz2017,Popovic2020} in which an informative path planning framework is presented for 2D terrain mapping and environmental monitoring. We adopt the similar framework and adapt it for 3D surface active information gathering. In particular, we use manifold Gaussian processes (mGPs) with geodesic kernel functions to map the surface information fields which encode their spatial correlations. For data measurement and map update, a probabilistic sensor (camera) model with limited field-of-view (FoV) is considered. Based on the map, a continuous trajectory is optimized by maximizing an information acquisition metric in a receding horizon fashion. The planned trajectory is collision-free and respects the UAV dynamics. 

The main contributions of this work are:
\begin{itemize}
    \item An online informative path planning approach for active information gathering on three-dimensional surfaces. 
    \item The use of manifold Gaussian processes with geodesic kernel functions for mapping surface information fields. 
\end{itemize}

We show that our online planning approach can generate more informative paths than the coverage planner. Moreover, by taking account of spatial correlations of the information field using manifold Gaussian processes, the information gathering efficiency is significantly improved.

%% file: secs/02_related_work.tex
%%%%%%%%%%%%%%%%%%%%%%%%%%%%%%%%%%%%%%%%%%%%%%%%%%%%%%%%%%%%%%%%%%%%%%%%%%%%%%%%
\section{Related Work}\label{sec:relatedWork}

In path planning for inspection of a given surface manifold, most existing works formulate it as an offline coverage planning problem which tries to find a path that can provide full coverage of the surface \cite{galceran2013survey}. 
Early works including \cite{Atkar2001,Acar2002} divide non-planar surfaces into regions using exact cellular decompositions that are then covered in a spiral sweeping pattern. Alternatively, \cite{Hover2012} presents a two-step process for inspection path planning which first constructs a set of viewpoints and then finds a short path connecting them by solving a traveling salesman problem (TSP). A similar approach is developed in \cite{Bircher2016} in which the two steps are iteratively performed to improve the resulting path. By employing an iterative strategy with re-meshing techniques, \cite{Alexis2015} proposes a path planning framework that provides uniform coverage of 3D structures. Besides those optimization-based approaches, several sampling-based coverage planning methods \cite{Englot2012,Papadopoulos2013,bircher2017incremental} are presented by exploring the configuration space with a random tree. In contrast to these offline path planning works, there are also several online planning approaches for surface inspection. \cite{Pito1999} adapts the next-best-view (NBV) planner \cite{Connolly1985} to automated surface acquisition by enforcing the surface portions scanning and overlap constraints. Recently \cite{Bircher2018} presents a receding horizon NBV planner for inspection of a given surface manifold. Alternatively, \cite{Yoder2016} proposes an incremental path planning algorithm using surface frontiers to guide the observation of a surface. 

While the aforementioned path planning approaches can generate offline coverage paths or online NBV paths, they all inherently assume that the information on the surface is uniformly distributed, hence ignoring potential spatial correlations of the information field. To model such information spatial correlations, Gaussian processes (GPs) \cite{williams2006gaussian} have been used as a popular mapping method and successfully applied in terrain mapping \cite{Popovic2020}, target search \cite{Meera2019} and environmental monitoring \cite{Hitz2017}. However, the normal GP formulation is limited to 2D terrains or 3D Euclidean space. To facilitate mapping 3D surfaces, Gaussian process implicit surfaces (GPISs) \cite{Williams2007} have been used for surface reconstruction \cite{Kim2015GP,Lee2019,Wu2020}, object shape estimation \cite{Dragiev2011}, and pipeline thickness mapping \cite{Vidal-Calleja2014}. The key idea is to represent the surface using a function that specifies whether a point in space is on the surface, outside the surface, or inside the surface. However, GPISs are limited to modelling surface geometry and cannot be directly applied to general surface information fields (e.g. temperature distribution over a surface). Recently, manifold Gaussian processes (mGPs) \cite{Jayasumana2015} have been developed to map information fields to complex domains and surfaces with heat kernels \cite{Niu2019}, generalized Mat\'ern kernels \cite{Borovitskiy2020}, and geodesic Gaussian kernels \cite{Sugiyama2008,DelCastillo2015}. In this paper we use manifold Gaussian processes with geodesic kernel functions to map the surface information fields and plan informative paths based on the map.

%% file: secs/03_formulation.tex
%%%%%%%%%%%%%%%%%%%%%%%%%%%%%%%%%%%%%%%%%%%%%%%%%%%%%%%%%%%%%%%%%%%%%%%%%%%%%%%%
\section{Problem Statement}\label{sec:formulation}
Given a 3D surface model with known geometry described by a mesh that can be obtained from CAD software or previous mapping missions, our goal is to plan a continuous dynamically-feasible trajectory for a UAV that maximizes its information gathering about the surface within a flight time budget. 
Formally, an informative path planning (IPP) problem is formulated as follows,
\begin{subequations}\label{eq:IPP}
	\begin{alignat}{2}
		\cT^* = &\argmax_{\cT \in \mathbb{T}}\frac{\tn{I}(\tn{MEASURE}(\cT))}{\tn{TIME}(\cT)} \\
        \text{s.t.}~~ &\tn{TIME}(\cT) \leq B, \qquad \label{subeq:budget_constraint} \\
        &\tn{COLLISION}(\cT) = \emptyset, \label{subeq:collision_avoidance}
	\end{alignat}
\end{subequations}
where $\cT^*$ is the planned optimal trajectory, $\mathbb{T}$ indicates the set of all continuous feasible trajectories, and $B$ denotes the specified flight time budget. The function $\tn{MEASURE}(\cdot)$ obtains a finite sequence of measurements along the trajectory and $\tn{I}(\cdot)$ quantifies the gathered information from those measurements. The function $\tn{TIME}(\cdot)$ and $\tn{COLLISION}(\cdot)$ return total flight duration and collision sections of the trajectory, respectively. Hence, by enforcing constraints Eq. (\ref{subeq:budget_constraint}) and (\ref{subeq:collision_avoidance}), the planned trajectory should respect the time budget and be collision-free in the environment.

%% file: secs/04_method_1.tex
%%%%%%%%%%%%%%%%%%%%%%%%%%%%%%%%%%%%%%%%%%%%%%%%%%%%%%%%%%%%%%%%%%%%%%%%%%%%%%%%
\section{Method} \label{sec:method}
In this section we present our surface mapping and informative path planning approach. The approach builds upon the previous work in \cite{Popovic2020} for 2D terrain mapping. We adapt it to 3D surface active information gathering by using manifold Gaussian processes (mGPs) for information field mapping and planning 3D collision-free viewpoint trajectories based on the map. 

\subsection{Method Overview}
\begin{figure*}[t]
	\centering
	\includegraphics[width=0.7\linewidth]{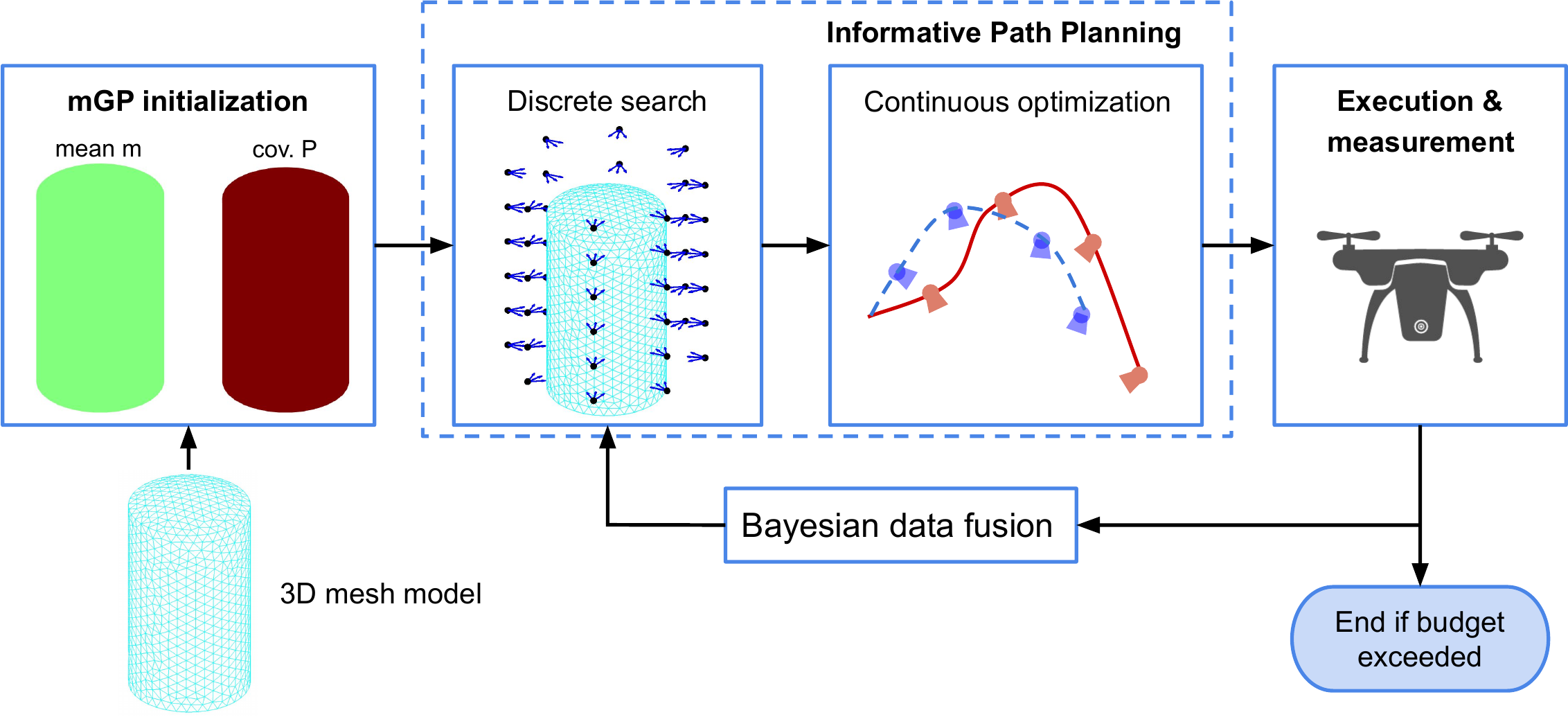}
	\caption{Overview of the proposed informative path planning approach for active information gathering on a 3D surface. }
	\label{fig:method_overview}
\end{figure*}

Fig. \ref{fig:method_overview} depicts an overview of the proposed approach. Given a mesh model of the 3D surface to be inspected, we first initialize an information map, including mean and covariance data, using a manifold Gaussian process. Based on the map, informative path planning is performed to generate a local continuous trajectory which is then executed by the UAV to take measurements and update the map through data fusion. The informative path planning module includes two stages: a) finding a finite sequence of viewpoints from a predefined library via discrete search, followed by b) optimizing a continuous trajectory that maximizes information acquisition using the discrete search result as an initial guess. Finally, the path planning, measurement taking, and map update are run online in a closed-loop manner until the UAV's flight time exceeds the specified budget. In the following, we will describe each module of the approach in detail.  

\subsection{Surface Information Mapping} \label{subsec:surface_mapping}
The information field on the surface is assumed to be a continuous function defined on a Riemannian manifold $f:\cX\ra\R$. Let $x \in \cX \subset \R^3$ be some location on the surface. We use an mGP to map the information field $f \sim \cG\cP(\mu, k)$ with mean function $\mu(x)$ and covariance function (kernel) $k(x, x^\prime)$, which encodes the correlation of field values at the two locations $x$ and $x^\prime$. In this paper, we use the geodesic Mat\'ern 3/2 kernel function to model such spatial correlations on a surface manifold~\cite{Sugiyama2008,DelCastillo2015}:
\begin{equation}
    k(x,x^\prime) = \sigma_f^2\left(1+\frac{\sqrt{3}d_g(x,x^\prime)}{l}\right)\exp\left(\frac{-\sqrt{3}d_g(x,x^\prime)}{l}\right),
\end{equation}
where $d_g(x,x^\prime)$ is the geodesic distance between $x$ and $x^\prime$ on $\cX$.

We assume the surface is discretized and represented by a mesh with $n$ triangle facets, which can be obtained from CAD software. The field value of each facet is assumed to be the value of the triangle's center. Given a set of noisy measurements $\vy = [y_1, \dots, y_m]^T$ at locations $\cX^*=[x_1,\dots,x_m]^m$, a posterior distribution of the field can be calculated using GP regression \cite{williams2006gaussian}:
\begin{equation}\label{eq:GP_regression}
    \begin{aligned}
        \mu &= \mu(\cX^*) + K_{\cX^*\cX}[K_{\cX}+\sigma^2\vI_n]^{-1}(\vy - \mu(\cX)), \\
        P &= K_{\cX^*} - K_{\cX^*\cX}[K_{\cX} + \sigma^2\vI_n]^{-1}K_{\cX^*\cX},
    \end{aligned}
\end{equation}
where $K_{\cX^*\cX} = k(x,x^\prime)|_{x\in\cX^*}^{x^\prime\in\cX}$, $K_{\cX^*} = k(x,x^\prime)|_{x\in\cX^*}^{x^\prime\in\cX^*}$ and $K_{\cX} = k(x,x^\prime)|_{x\in\cX}^{x^\prime\in\cX}$, $\sigma^2$ is a noise variance hyperparameter. 

We assume the UAV has a constant sensor measurement frequency. When new measurement data is taken by the UAV, we need to fuse it into the mGP field map. However, the regression in Eq. (\ref{eq:GP_regression}) is very computationally expensive \cite{williams2006gaussian}, particularly for large fields, which hinders its use in real-time applications. Hence, in this paper we utilize a Bayesian fusion technique \cite{Sun2015,Popovic2020} to perform sequential map updates as follows:
\begin{align}
    \mu^+ &= \mu^- + P^-H^T(HP^-H^T+R)^{-1}(\vy - H\mu^-), \\
    P^+ &= P^- - P^-H^T(HP^-H^T+R)^{-1}HP^-, \label{eq:cov_update}
\end{align}
where the superscripts $^-$ and $^+$ indicate variables before and after data fusion, respectively; $H\in\R^{m\times n}$ is the observation matrix which intrinsically selects part of the locations $x_1,\dots,x_m$ from all $n$ triangle facets that are observed through $\vy$; $R\in\R^{m\times m}$ is the observation noise matrix. 
% Note that the field covariance update in Eq. (\ref{eq:cov_update}) is only dependent on where the observations are taken and does not need to know the actual measurements. 

\subsection{Sensor Model} \label{subsec:sensor_model}
\begin{figure}[t]
    \centering
    \captionsetup[subfigure]{position=b}
    \begin{subfigure}{0.133\textwidth}
            \includegraphics[width=1.0\textwidth]{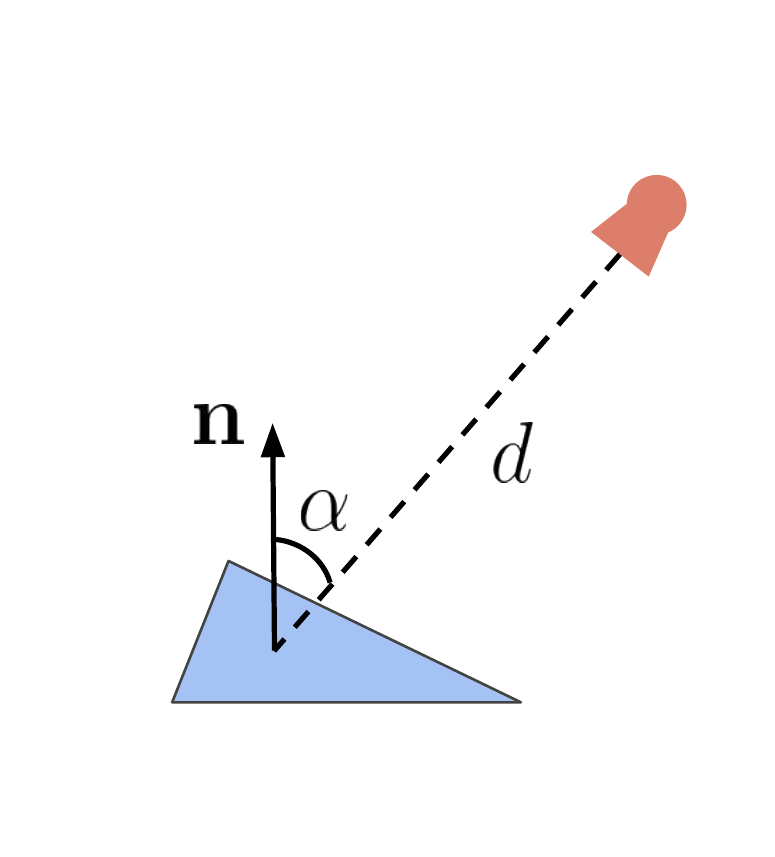}
	\caption{}
	\label{subfig:camera_model}
    \end{subfigure}
    \captionsetup[subfigure]{position=b}
    \begin{subfigure}{0.105\textwidth}
            \includegraphics[width=1.0\textwidth]{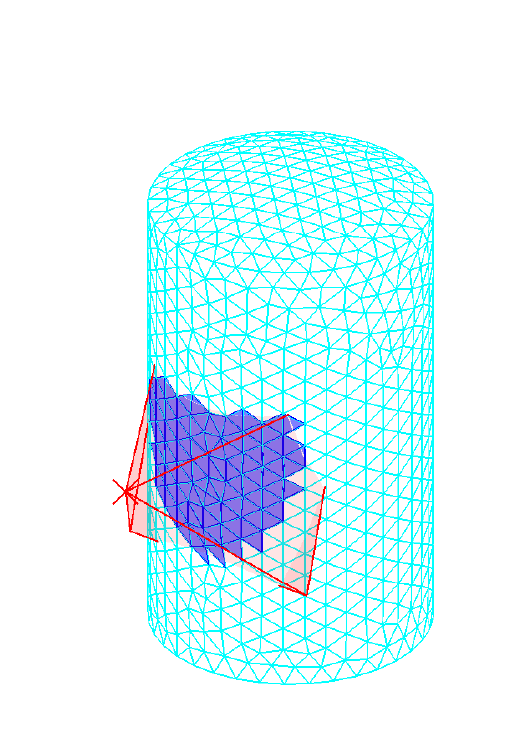}
	\caption{}
	\label{subfig:cylinder_camera}
    \end{subfigure}
    ~~~
    \captionsetup[subfigure]{position=b}
    \begin{subfigure}{0.11\textwidth}
            \includegraphics[width=1.0\textwidth]{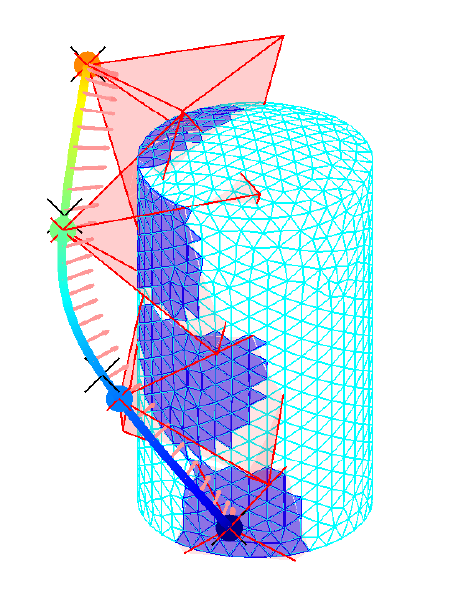}
	\caption{}
	\label{subfig:cylinder_path_one}
    \end{subfigure}
    ~~
    \caption{Sensor model of the camera with a limited FoV and a constant measurement frequency. (a) Illustration of the range $d$ and incidence angle $\alpha$ of a triangle facet with respect to the camera. (b) An example of triangle facets that are visible to the camera. (c) An example of a path segment along which measurements are taken by the camera (path control waypoints as black crosses, measurement viewpoints as filled circles, UAV orientation as the light pink arrow, and visible triangle facets in blue). }
    \label{fig:sensor_model}%
\end{figure}

We consider the UAV carries a camera with fixed orientation relative to the platform and a limited FoV. For a triangle facet on the surface to be considered visible by the camera, it must satisfy the following conditions: i) its center is in the camera's FoV; ii) its center's distance to the camera $d$ is within a valid range $d \in [d_{\tn{min}}, d_{\tn{max}}]$; iii) its incidence angle with respect to the camera $\alpha$ should be smaller than a maximal valid angle $\alpha \leq \alpha_{\tn{max}}$; and iv) it is not occluded by other parts of the surface. Fig. \ref{fig:sensor_model}(a)-(c) show an illustrative example of a camera inspecting the surface of a cylinder storage tank. 

In addition to the geometry model, we also consider a range-dependent Gaussian sensor noise model as in \cite{Popovic2020}. Typically, when the camera is closer to the observation location, its measurement uncertainty is smaller, for example the thermal camera \cite{Kennedy1993}. Formally, the measurement noise variance $\sigma_i^2$ of the camera with respect to the observation location $x_i$ is modeled as:
\begin{equation}
    \sigma_i^2 = a(1-e^{-bd_i}),
\end{equation}
where $d_i$ is the distance between the triangle facet center and the camera, $a$ and $b$ are positive constant parameters. 

\subsection{Online Path Planning} \label{subsec:path_planning}
We now present our path planning method based on the mGP surface map and the probabilistic sensor model. The method builds upon the framework developed in \cite{Hitz2017,Popovic2020} for 2D terrain mapping. We extend it to 3D surface inspection by further planning yaw trajectories of the UAV and enabling collision avoidance. 

\subsubsection{Trajectory parametrization}
We represent the continuous trajectory $\cT$ of the UAV in a polynomial form parameterized by a sequence of $N$ control waypoints $\cC = \{\vc_1,\dots,\vc_N\}$ to be visited. Each control waypoint includes a position $\vp$ and yaw angle $\psi$ according to the differential flatness property of UAVs \cite{Mellinger2011}. The polynomial trajectory connects these control points using $N-1$ $k$-order spline segments for minimum-snap dynamics given a reference speed and acceleration as shown in \cite{Richter2016}. Measurements are then taken by the UAV along the trajectory with a constant frequency, as illustrated in Fig. \ref{fig:sensor_model}(c).
% \ref{subfig:cylinder_path_one}. 

\subsubsection{Discrete waypoint search}
As depicted in Fig. \ref{fig:method_overview}, in each planning stage, to obtain the optimal control waypoints, we first perform a sequential greedy search in a predefined viewpoints library $\cL$. The algorithm is shown in Alg.~\ref{Alg:grid_search}. Given the current viewpoint $\vc_{\tn{prev}}$, the next best viewpoint $\vc^*$ is chosen from the library $\cL$ such that it has the highest information gathering efficiency (Line 4) in which the information gain is defined as \cite{Popovic2020},
\begin{equation}\label{eq:info_gain}
    \tn{I}(\vc) = \tn{Tr}(P^-) - \tn{Tr}(P^+),
\end{equation}
where $\tn{Tr}(\cdot)$ is the trace of a matrix, $P^-$ and $P^+$ denote the surface map covariance before and after taking a measurement at $\vc$. Note that updating the map covariance (Line 7) to compute $P^+$ does not need the real measurement data. Instead, only the measurement viewpoint of the UAV is needed, as shown in Eq. (\ref{eq:cov_update}). This enables us to simulate the map covariance evolution during the planning stage before real trajectory execution. 

\begin{algorithm}[t]
	\caption{ Sequential Greedy Waypoint Search}
    \label{Alg:grid_search}
	\begin{algorithmic}[1]
    %	\Statex ------------------  ------------------
        \Statex \tb{Input:} Current surface map $\cM$, current viewpoint $\vc_1$, 
        \Statex ~~~number of waypoints $N$, viewpoints library $\cL$
        \Statex \tb{Ouput:} A sequence of waypoints $\cC =\{\vc_1,\dots,\vc_N\}$
        \State $\cM^\prime \la \cM$
        \State $\cC \la \emptyset$; $\vc_{\tn{prev}} \la \vc_1$
        \While {$\abs{\cC} \leq N$}
            \State $\vc^* = \argmax_{\vc\in\cL}\frac{\tn{I}(\vc)}{\tn{TIME}(\vc_{\tn{prev}},\vc)}$
            \State $\cC = \cC \cup \vc^*$
            \State $\vc_{\tn{prev}} \la \vc^*$
            \State $\cM^\prime \la \tn{UPDATE\_COVARIANCE}(\cM^\prime, \vc^*)$ 
        \EndWhile
	\end{algorithmic}
\end{algorithm}

\subsubsection{Continuous optimization}
Using the sequential greedy search result as an initial guess, we further perform a continuous optimization to obtain the optimal solution. The decision variables of the optimization problem are the control waypoints $\cC$. We compute the objective function by first connecting these waypoints using a polynomial trajectory along which measurement viewpoints are found and used to update the map covariance. Then the objective to be maximized is defined as the time-averaged information gain:
\begin{equation}
    U_{\tn{info}} = \frac{\tn{I}(\tn{MEASURE}(\cT))}{\tn{TIME}(\cT)},
\end{equation}
where the gain function $\tn{I}(\cdot)$ is the same as in Eq. (\ref{eq:info_gain}) defined as the map covariance uncertainty reduction.

\subsubsection{Collision avoidance}\label{subsubsec:coll}
The planned trajectory must be collision-free with the structure to be inspected. To achieve that, we first build a 3D Euclidean Signed Distance Field (ESDF) based on the given surface mesh, which returns the Euclidean distance at each point in the space to its nearest obstacle. Using the ESDF, during the greedy waypoint search stage, we add a constraint that the next viewpoint $\vc^*$ must be in line of sight (LoS) with the current one $\vc_{\tn{prev}}$, which can provide preliminary collision avoidance for the following continuous optimization. The LoS detection is achieved by sampling positions from the line segment connecting the two viewpoints and checking if any sampled position is in collision with the mesh according to its ESDF value. Besides, on the continuous optimization stage, we add one more objective to penalize collisions:
\begin{equation}
    U_{\tn{coll}} = w_{\tn{coll}}\sum_{\vp \in \cT}g(\vp),
\end{equation}
where $\vp$ are sampled positions from the polynomial trajectory, $w_{\tn{coll}}$ is the weight coefficient and,
\begin{equation}
    g(\vp) = 
    \begin{cases}
        0, &\tn{if}~\tn{ESDF}(\vp) \geq r, \\
        -1, &\tn{otherwise},
    \end{cases}
\end{equation}
in which $r$ is the radius of the UAV. Hence, the final objective function to be maximized is,
\begin{equation}
    U = U_{\tn{info}} + U_{\tn{coll}}.
\end{equation}

\subsection{Discussion}

\subsubsection{Yaw optimization}
In Section \ref{subsec:path_planning}, the yaw angles in the control waypoints are optimized as one of the decision variables together with positions. In practice this would require a larger number of iterations to find the optimal solution, especially for a complex surface. An alternative way to handle the issue is to pre-compute a ``best yaw angle library'' which returns the yaw angle at each point in the space that can see as many triangle facets of the surface as possible. Then during the path planning stage, only positions of the control waypoints are optimized while the corresponding yaw angles are found from the pre-computed library. Although such yaw angles are sub-optimal, the resulting performance does not deteriorate much while the computation efficiency is substantially improved. 

\subsubsection{Occlusion checking}
Visibility checking for all triangle facets is performed during planning to construct the observation matrix $H$ given a camera viewpoint. As described in Section \ref{subsec:sensor_model}, a triangle facet determined to be visible by the camera must satisfy four conditions, for which occlusion checking is performed by determining if the triangle center is in LoS of the camera. This can be achieved using the same sampling technique as described in Section \ref{subsubsec:coll} for collision avoidance. However, practically such occlusion checking may be very time-consuming and it is within the optimization iterations thus leading to a very heavy computation burden. In this paper we assume the surface to be inspected is convex. Hence, occlusion checking can be ignored \cite{Adams2014}.

%% file: secs/06_results.tex
%%%%%%%%%%%%%%%%%%%%%%%%%%%%%%%%%%%%%%%%%%%%%%%%%%%%%%%%%%%%%%%%%%%%%%%%%%%%%%%%
\section{Results}\label{sec:results}
In this section we validate the proposed online informative path planing method in simulation by comparing it to the coverage path planner. We evaluate the effects of taking information spatial correlations into account when mapping and planning. We also present a complex surface inspection case in which the surface temperature of a Boeing-747 airplane is mapped by a UAV.

\begin{figure}[t]
    \centering
    \captionsetup[subfigure]{position=b}
    \begin{subfigure}{0.23\textwidth}
            \includegraphics[width=1.0\textwidth]{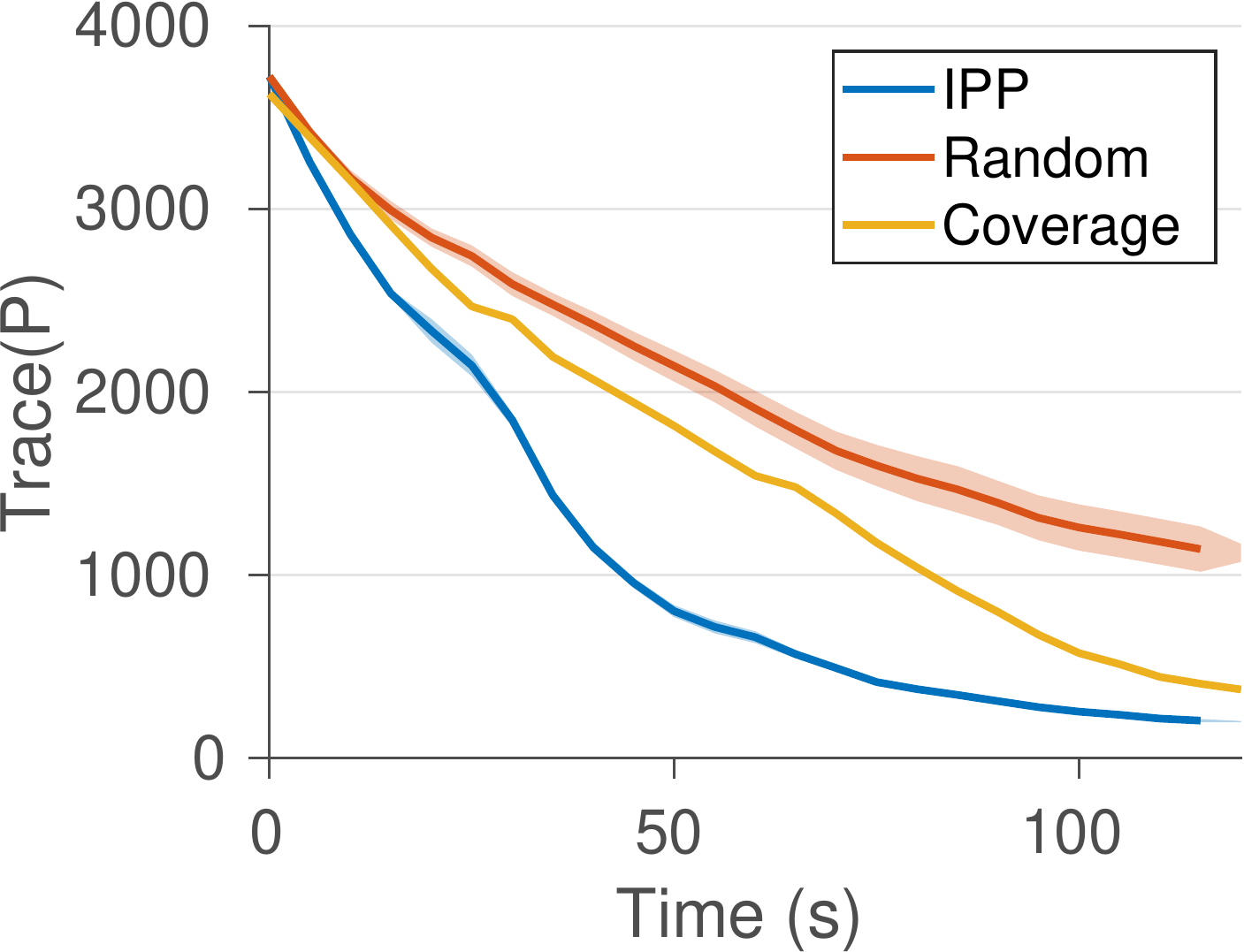}
	\caption{}
	\label{subfig:method_P}
    \end{subfigure}
    \captionsetup[subfigure]{position=b}
    \begin{subfigure}{0.23\textwidth}
            \includegraphics[width=1.0\textwidth]{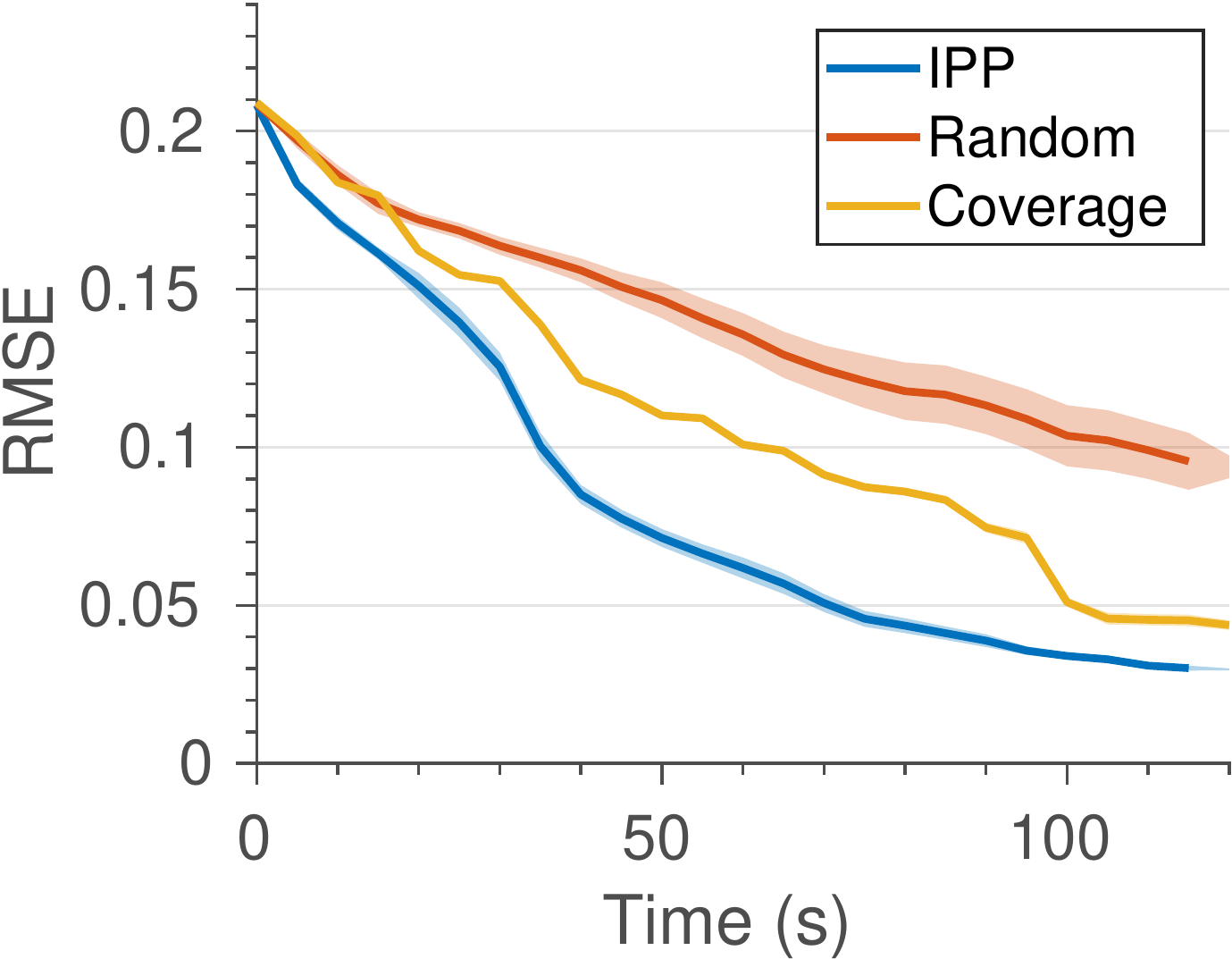}
	\caption{}
	\label{subfig:method_RMSE}
    \end{subfigure}
    \caption{Comparisons of the proposed informative path planner (IPP) to the coverage path planner and a random inspection strategy. The solid lines depict means of 30 trials for each method and the thin shaded regions indicate 95\% confidence bounds. (a) Trace of the map covariance. (b) Root mean squared error (RMSE) of the inspection results with respect to the ground truth.}
    \label{fig:method_eval}%
\end{figure}

\begin{figure}[t]
    \centering
    \captionsetup[subfigure]{position=b}
    \begin{subfigure}{0.16\textwidth}
            \includegraphics[width=1.0\textwidth]{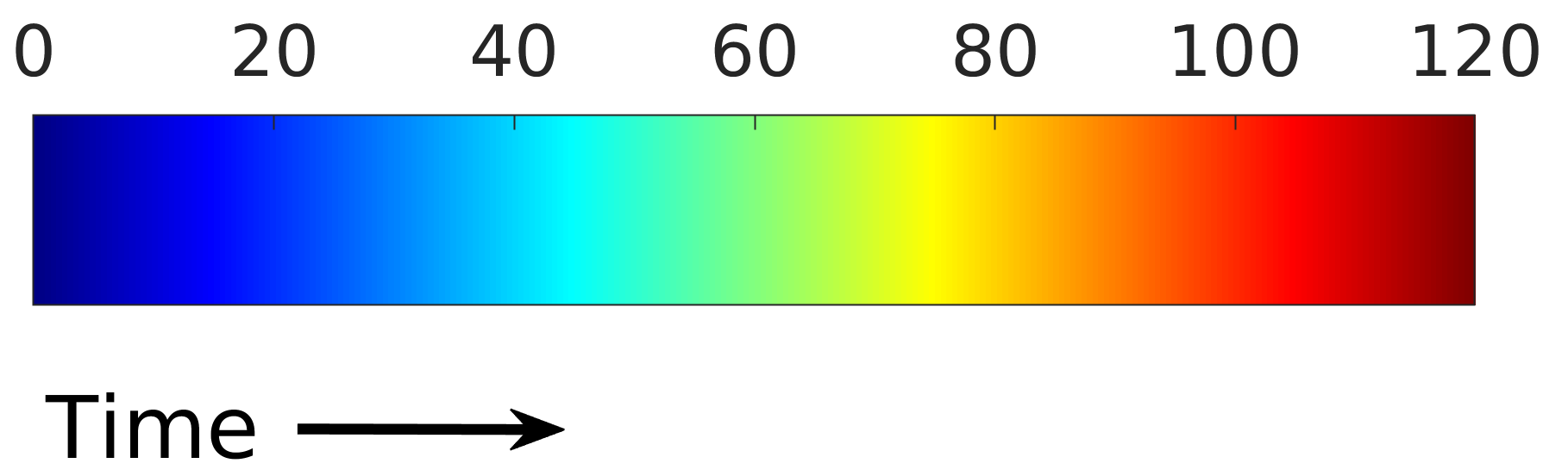}
	\label{subfig:cylinder_time_bar}
    \end{subfigure}
    \captionsetup[subfigure]{position=b}
    \begin{subfigure}{0.15\textwidth}
            \includegraphics[width=1.0\textwidth]{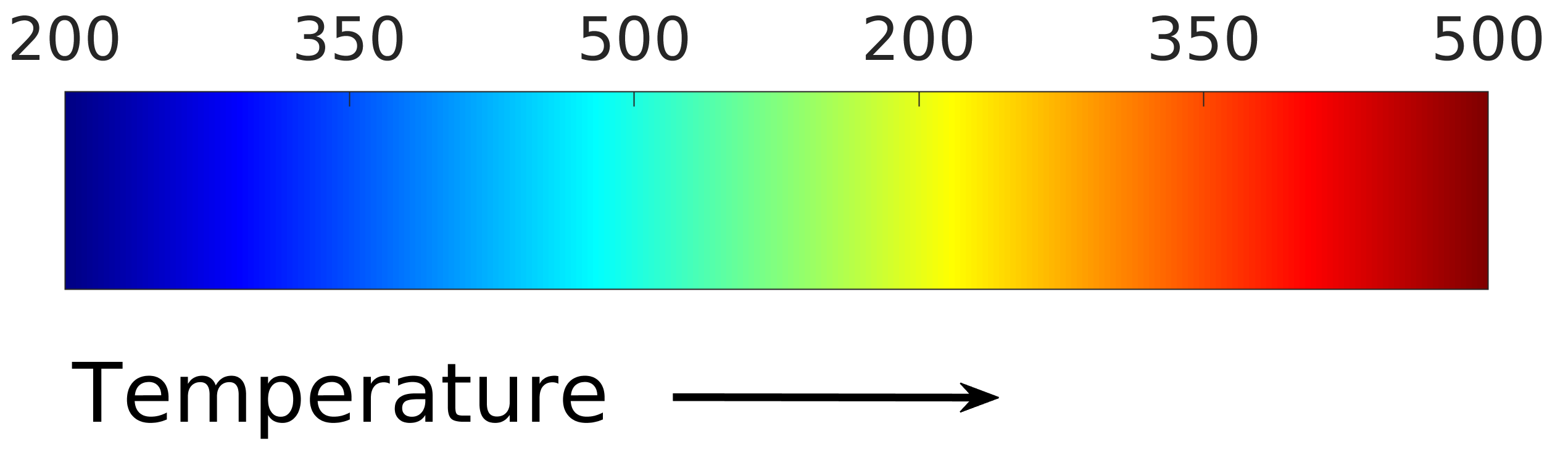}
	\label{subfig:cylinder_temperature_bar}
    \end{subfigure}
    \\
    \captionsetup[subfigure]{position=b}
    \begin{subfigure}{0.08\textwidth}
            \includegraphics[width=1.0\textwidth]{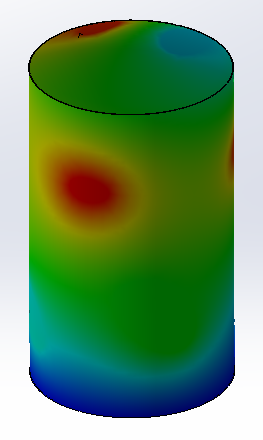}
	\caption{}
	\label{subfig:cylinder_ground_truth}
    \end{subfigure}
    \captionsetup[subfigure]{position=b}
    \begin{subfigure}{0.13\textwidth}
            \includegraphics[width=1.0\textwidth]{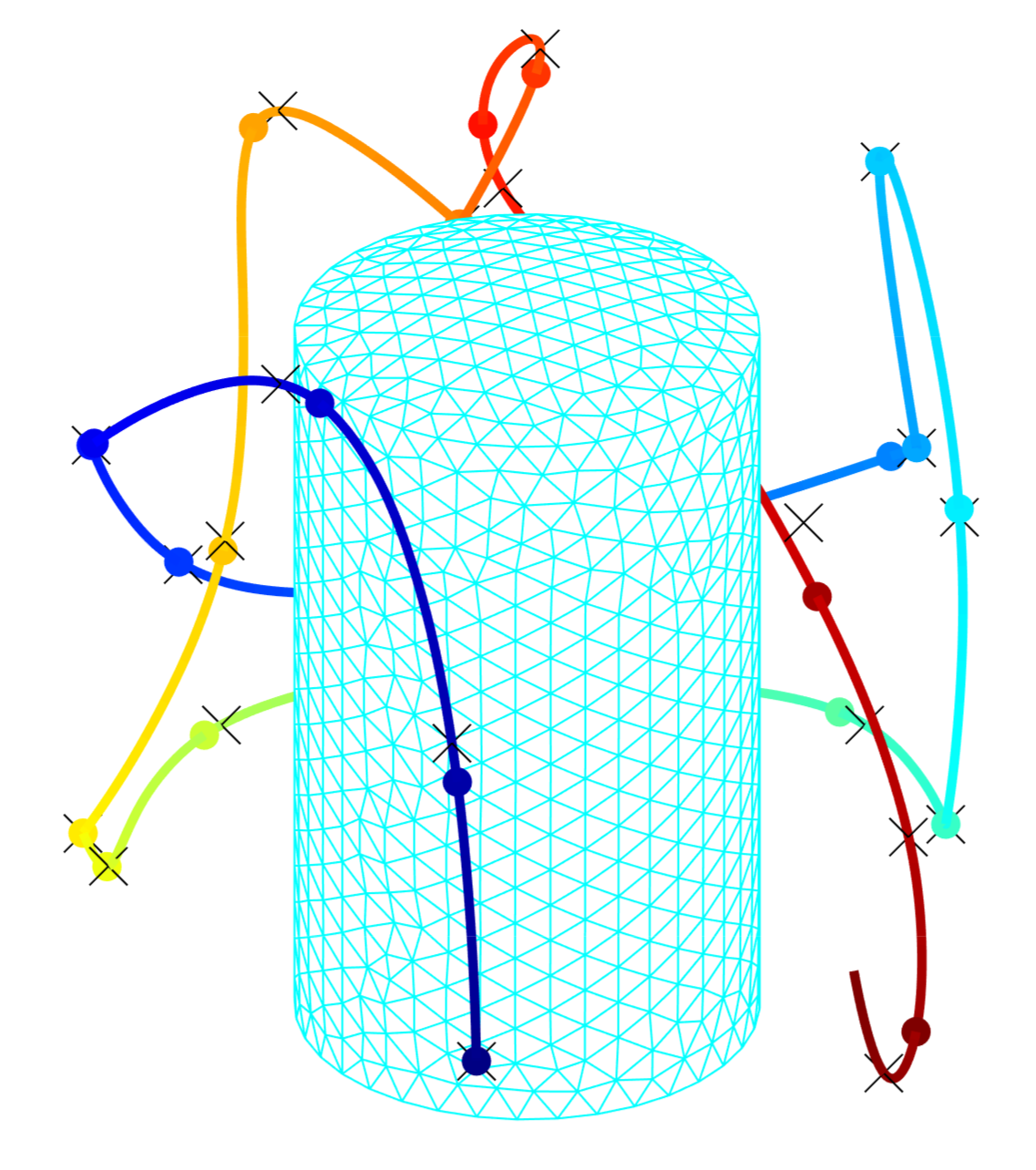}
	\caption{}
	\label{subfig:cylinder_path}
    \end{subfigure}
    \captionsetup[subfigure]{position=b}
    \begin{subfigure}{0.083\textwidth}
            \includegraphics[width=1.0\textwidth]{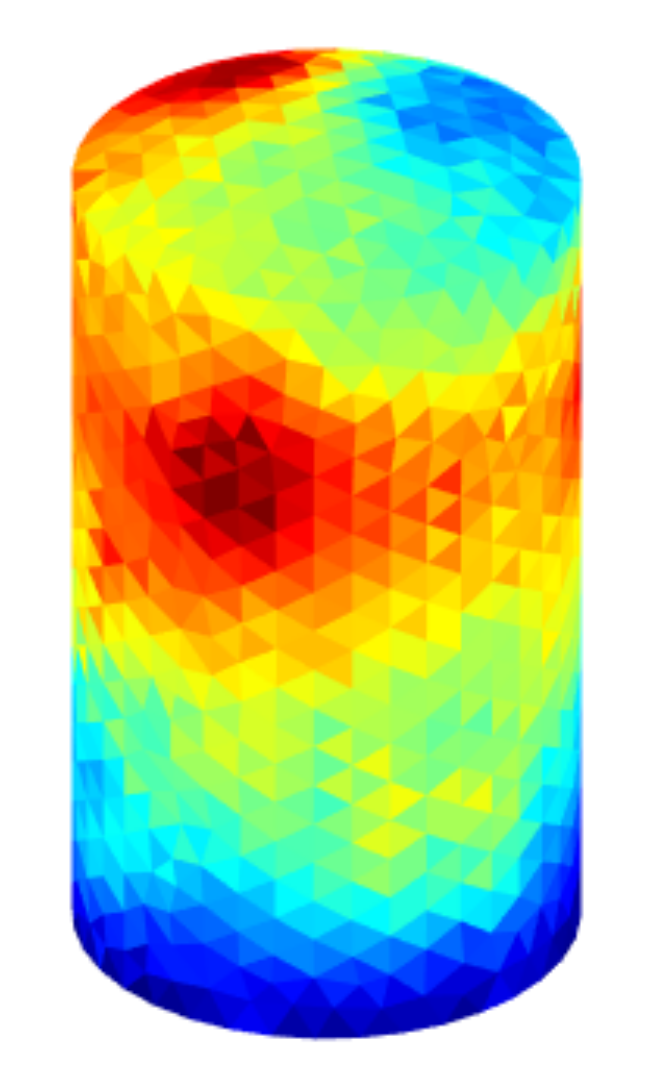}
	\caption{}
	\label{subfig:cylinder_map}
    \end{subfigure}
    \caption{Simulation results of our proposed IPP method for the cylinder storage tank inspection task. (a) A ground-truth temperature distribution simulated in SolidWorks 2019. (b) The inspection trajectory. (c) The inspection mapping results. Note that the color bars used in (a) and (c) are not the same due to different software settings. }
    \label{fig:cylinder_inspection}%
\end{figure}

\subsection{Comparisons to Coverage Planner}\label{subsec:com_method}
We first evaluate our proposed method on inspecting the surface of a chemical storage tank. The tank is in a cylinder shape with radius 6 m, height 20 m, thickness 0.2 m and it has a top dome with height 1.2 m. We create a 3D model of the tank with material alloy steel in SolidWorks 2019 and simulate temperature distributions on its surface as ground truth using the Thermal Simulation Tool by adding different heat sources. A mesh of the tank surface is generated by Delaunay triangulation using the open-source Gmsh software \cite{Geuzaine2009}, which results in 1097 nodes and 2190 triangle facets. The UAV has a maximum speed of 4 m/s, acceleration of 3 m/s$^2$, yaw rate of 90$\degree$/s, radius of 0.6 m, and measurement frequency of 0.2 Hz. A front camera is mounted to the UAV with a fixed pitch angle of $15\degree$, FoV of $(60,60)\degree$, valid sensor range of $[2, 8]$ m, maximal valid sensor incidence angle of 70$\degree$. The parameters of the range-dependent sensor noise model are $a = 0.05$ and $b = 0.2$. During simulation, Gaussian measurement noise is simulated for data fusion based on the noise model. 
% The UAV is initially located at $(-7, -7, 4)$ m with a yaw angle of $45\degree$.
For polynomial trajectory generation of the UAV, the number of control waypoints is $N = 4$ and the polynomial order is $k = 12$. The total time budget for the mission is $B = 120$s. The global optimization algorithm CMA-ES \cite{Hansen2007} is used in the continuous optimization stage. 

We compare our proposed approach to: a) a coverage path planner; and b) a random inspection strategy. The coverage planner is implemented by computing a set of viewpoints that provides full coverage of the surface offline and then connecting them with a polynomial trajectory executed online in a spiral sweeping pattern. In addition, a random inspection strategy is also implemented in which the UAV randomly chooses $N$ viewpoints from the library and generates a polynomial trajectory accordingly to take measurements. 30 trials are run for each method. 

Fig. \ref{fig:method_eval} shows the comparison results, in which we quantify the trace of the map covariance $\tn{Tr}(P)$ and the root mean squared error (RMSE) of the inspection results with respect to the ground truth. It can been from the figure that our presented IPP method outperforms the coverage path planner and the random inspection strategy in both reducing the map uncertainty $\tn{Tr}(P)$ and inspection error RMSE. More precisely, the coverage path reduces the map uncertainty and error in a uniform manner while our planned informative path achieves much faster uncertainty and error reduction, particularly during the early stage of the inspection mission. This indicates that the planned informative path is more efficient in gathering information than the coverage path. Another interesting observation is that the random inspection strategy can achieve a reasonable performance at the beginning since most regions are not mapped. However, with time, its performance becomes worse since it may fly back to those observed regions while still leaving many other regions unobserved. Fig. \ref{fig:cylinder_inspection} shows the inspection path and mapping results of one example trial using our proposed IPP method. As shown in the figure, the UAV successfully maps the surface temperature of the cylinder storage tank within the given flight time budget. 

\begin{figure}[t]
    \centering
    \captionsetup[subfigure]{position=b}
    \begin{subfigure}{0.23\textwidth}
            \includegraphics[width=1.0\textwidth]{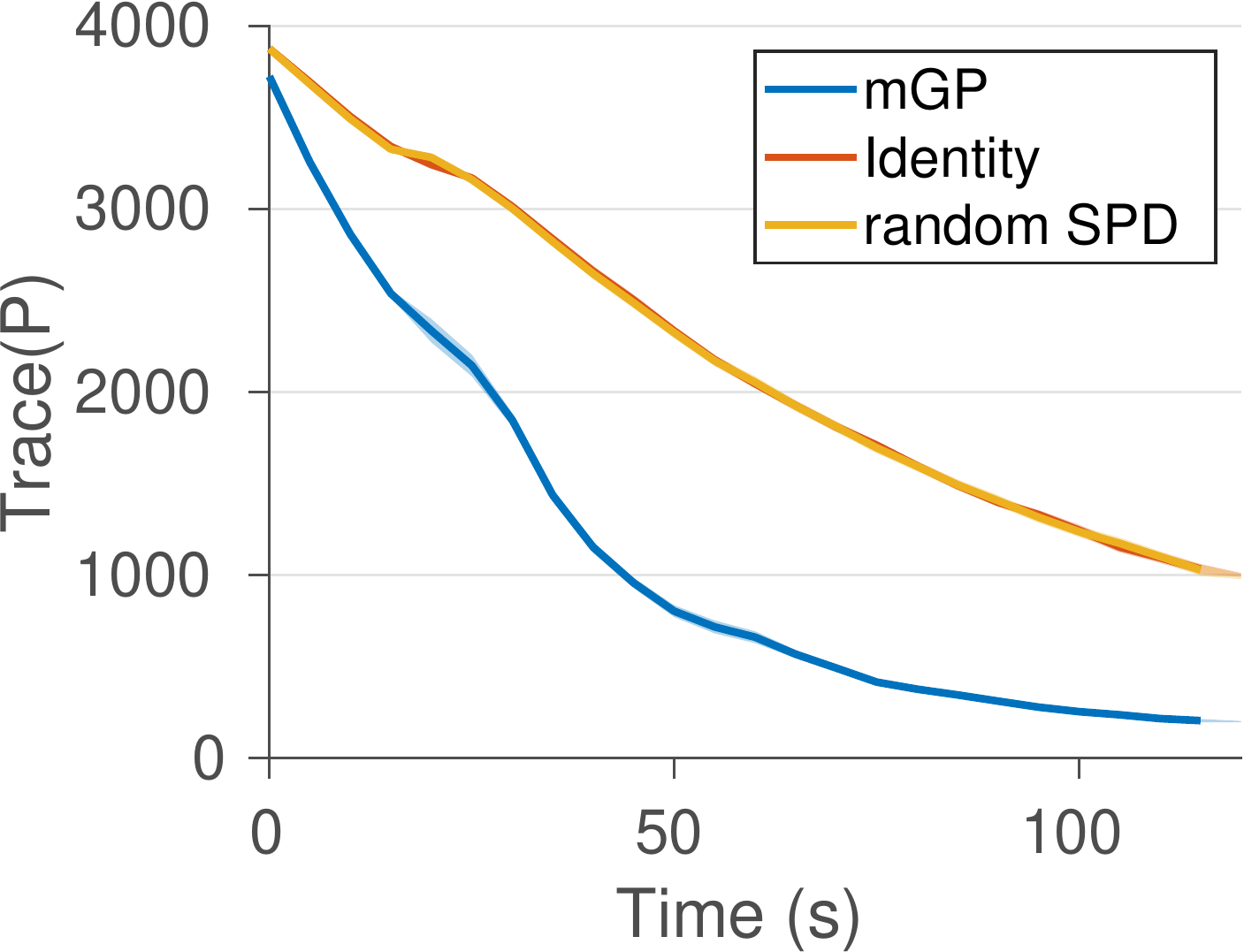}
	\caption{}
	\label{subfig:correlation_P}
    \end{subfigure}
    \captionsetup[subfigure]{position=b}
    \begin{subfigure}{0.23\textwidth}
            \includegraphics[width=1.0\textwidth]{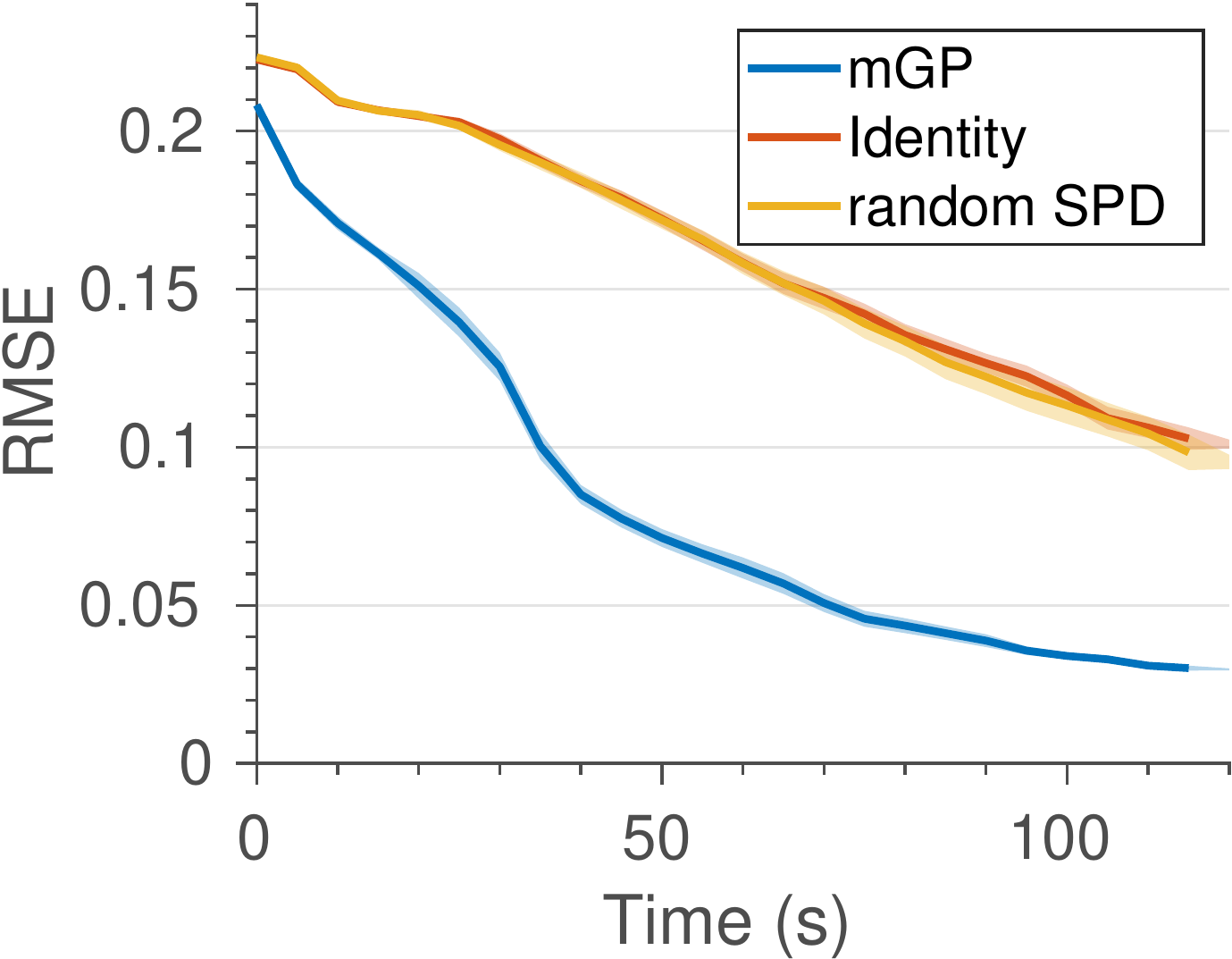}
	\caption{}
	\label{subfig:correlation_RMSE}
    \end{subfigure}
    \caption{Comparison results of different map covariance initialization methods: manifold Gaussian process (mGP), identity matrix, and random semi-positive definite matrix (SPD). 
    % For each method 30 trials have been run. 
    % The solid lines depict means of 30 trials for each method and the thin shaded regions indicate 95\% confidence bounds. (a) Trace of the map covariance. (b) Root mean square error (RMSE) of the inspection results with respect to the ground truth.
    }
    \label{fig:correlation_eval}%
\end{figure}

\begin{figure*}[t]
    \centering
    \captionsetup[subfigure]{position=b}
    \begin{subfigure}{0.24\textwidth}
            \includegraphics[width=1.0\textwidth]{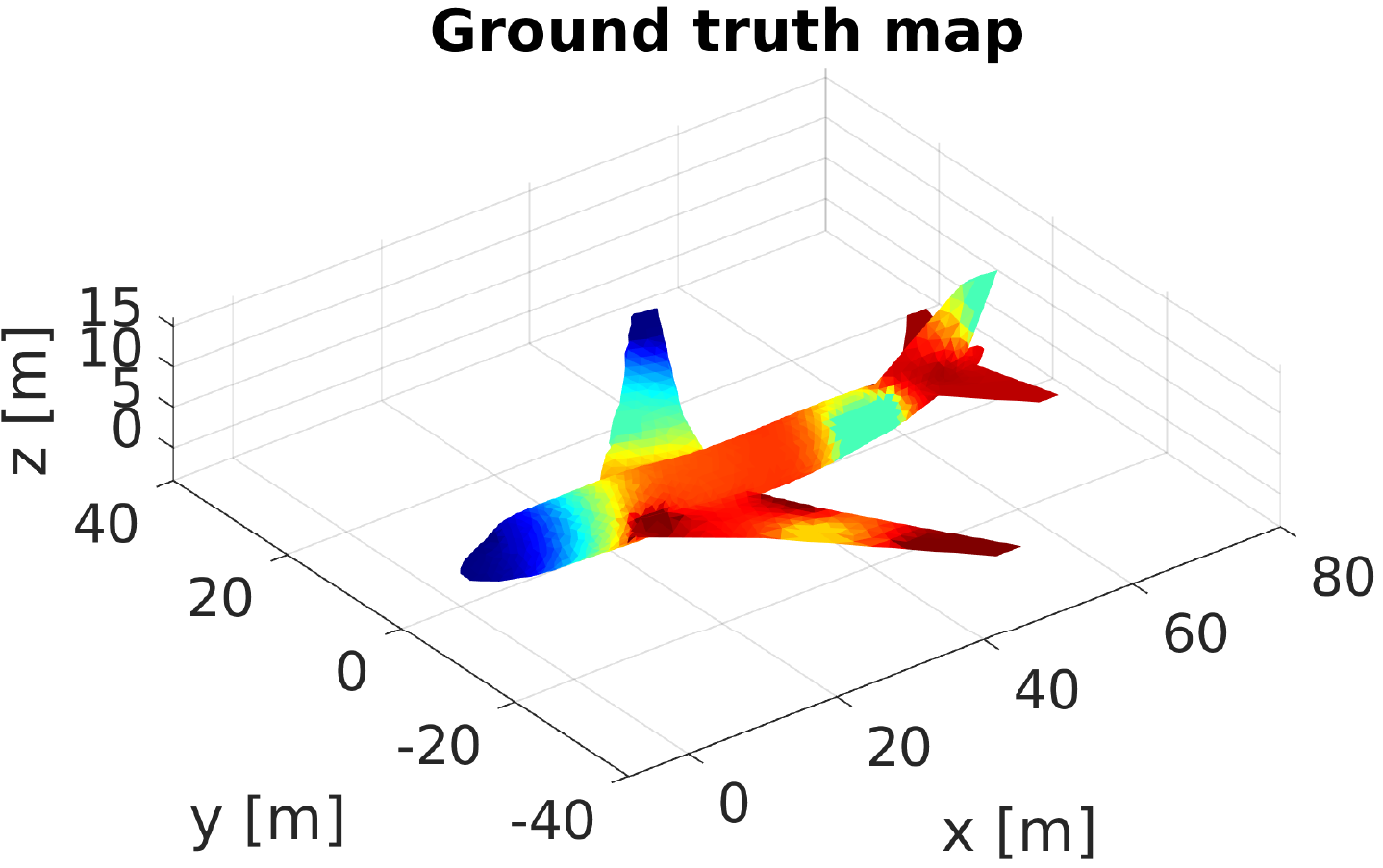}
	% \caption{}
	\label{subfig:boeing747_ground_truth}
    \end{subfigure}
    \captionsetup[subfigure]{position=b}
    \begin{subfigure}{0.24\textwidth}
            \includegraphics[width=1.0\textwidth]{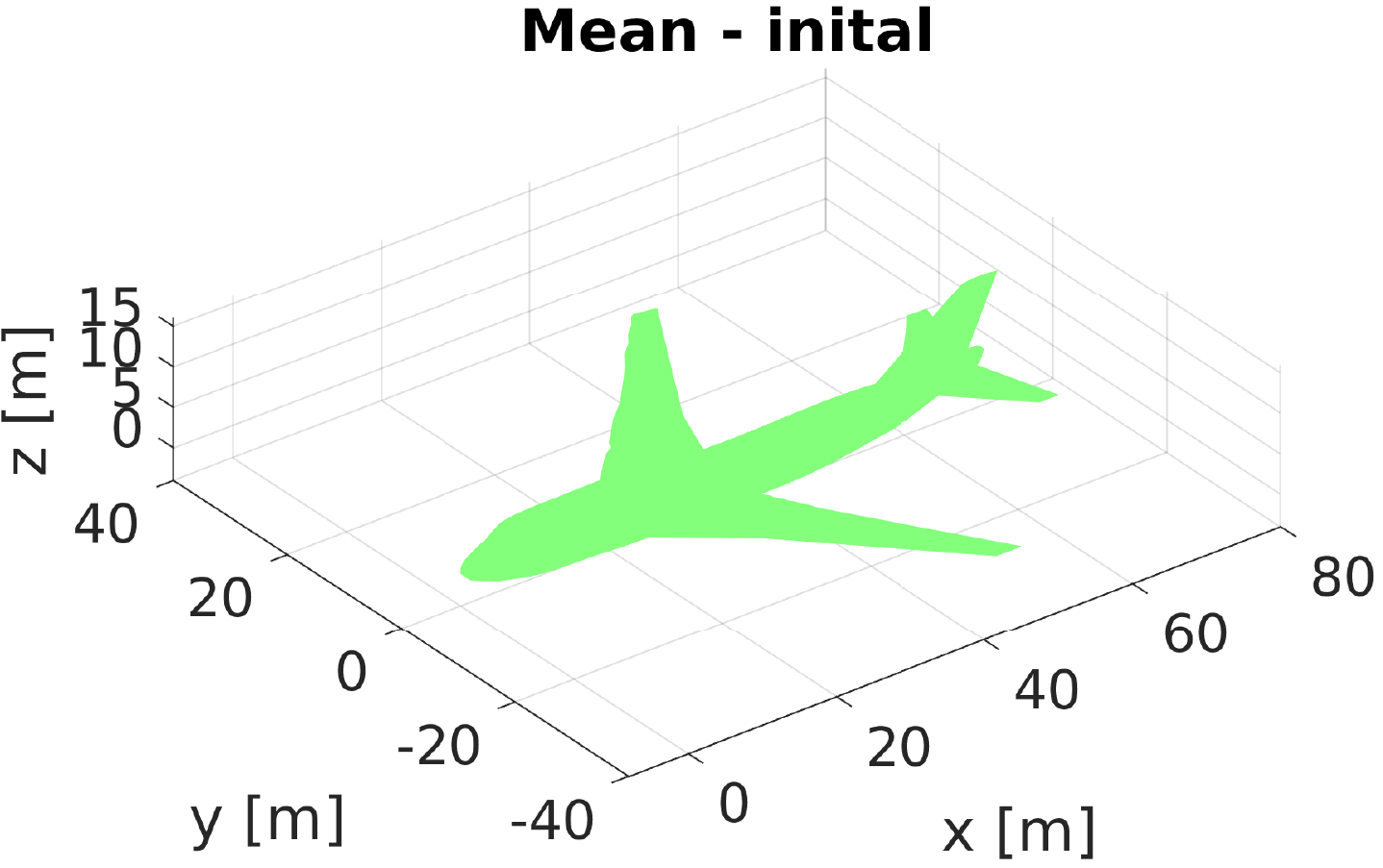}
	% \caption{}
	\label{subfig:boeing747_mean_initial}
    \end{subfigure}
    \captionsetup[subfigure]{position=b}
    \begin{subfigure}{0.24\textwidth}
            \includegraphics[width=1.0\textwidth]{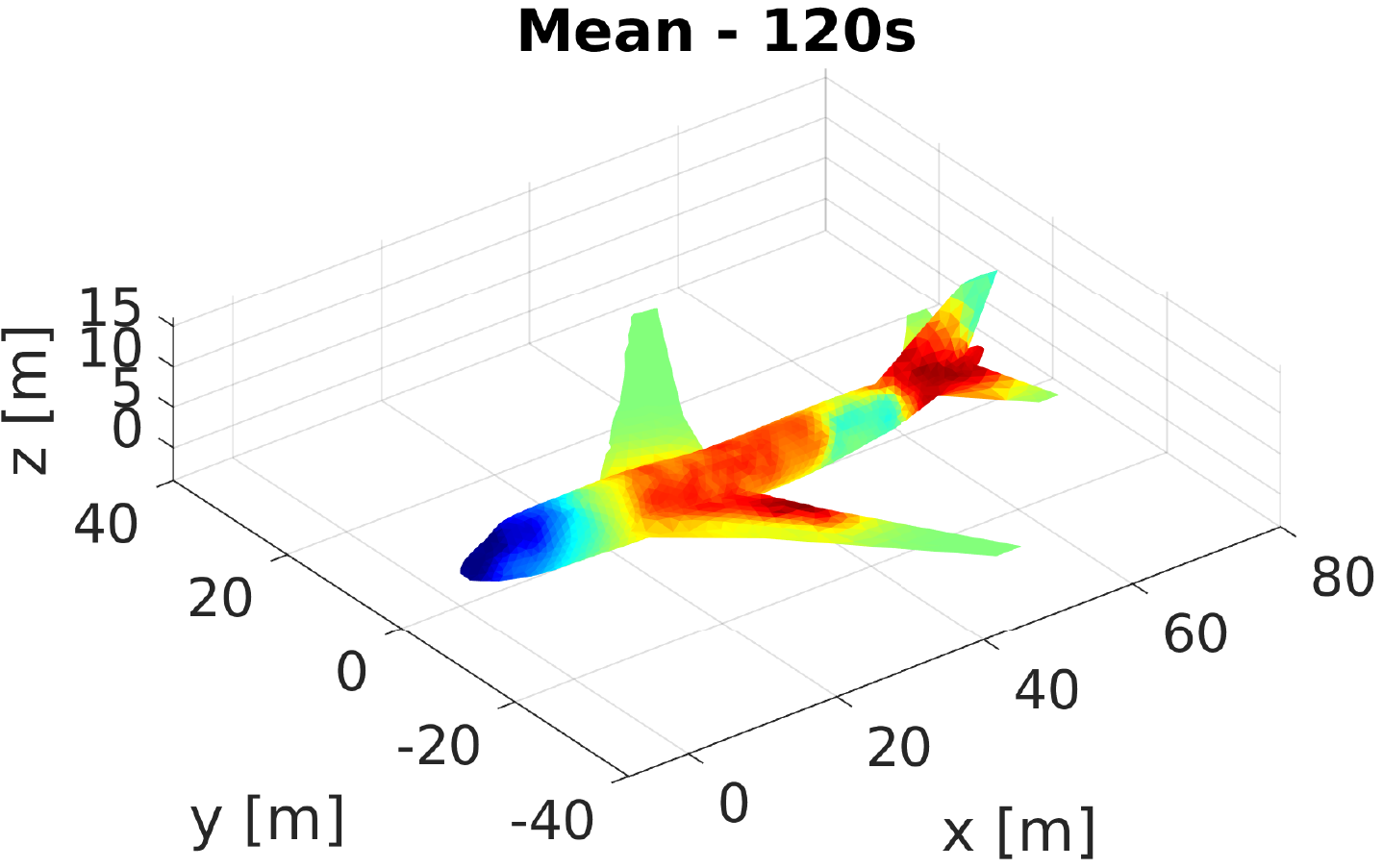}
	% \caption{}
	\label{subfig:boeing747_mean_120}
    \end{subfigure}
    \captionsetup[subfigure]{position=b}
    \begin{subfigure}{0.24\textwidth}
            \includegraphics[width=1.0\textwidth]{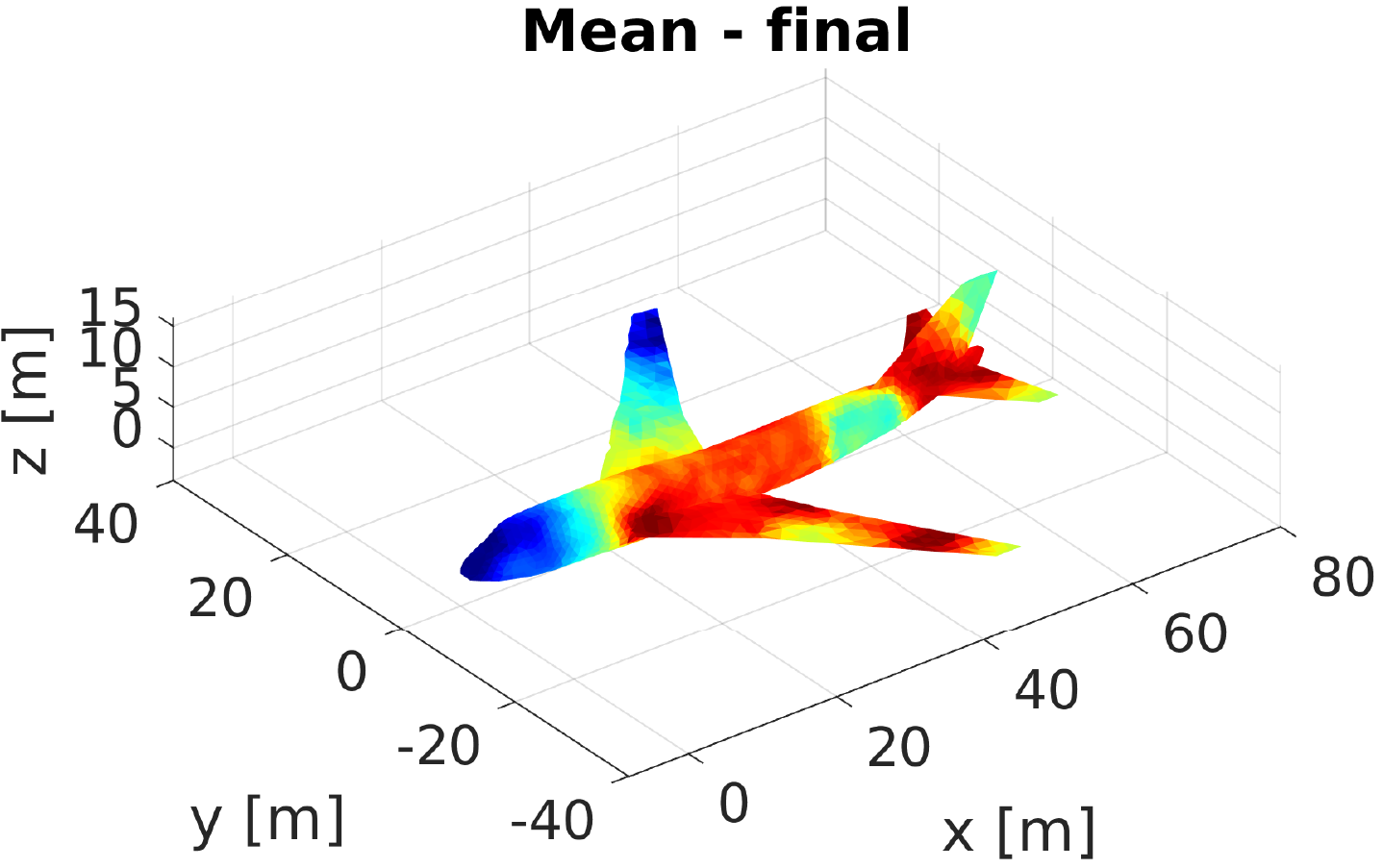}
	% \caption{}
	\label{subfig:boeing747_mean_final}
    \end{subfigure}
    \\
    \captionsetup[subfigure]{position=b}
    \begin{subfigure}{0.24\textwidth}
            \includegraphics[width=1.0\textwidth]{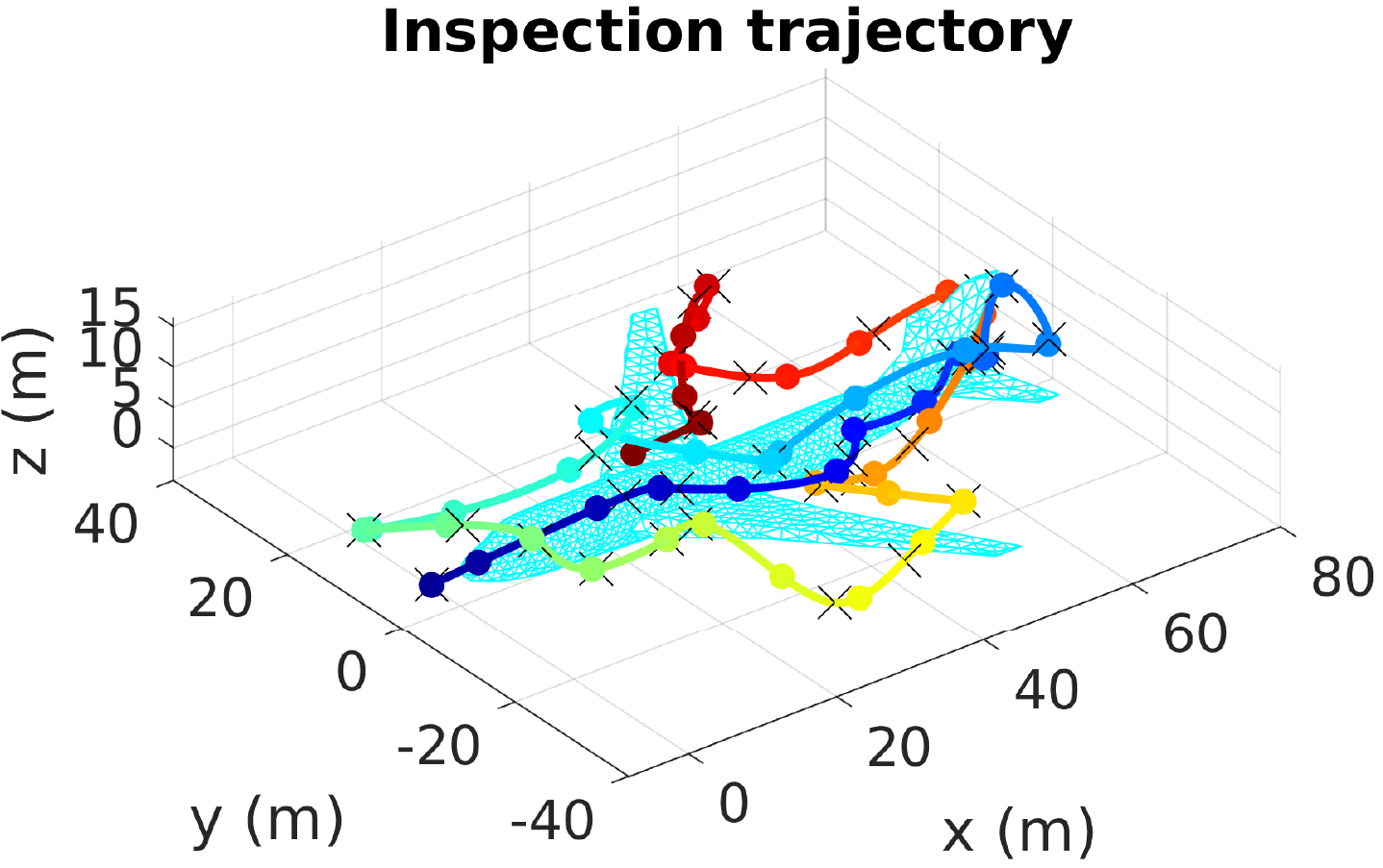}
	\caption{Ground truth map (top) and inspection trajectory (bottom).}
	\label{subfig:boeing747_trajectory}
    \end{subfigure}
    \captionsetup[subfigure]{position=b}
    \begin{subfigure}{0.24\textwidth}
            \includegraphics[width=1.0\textwidth]{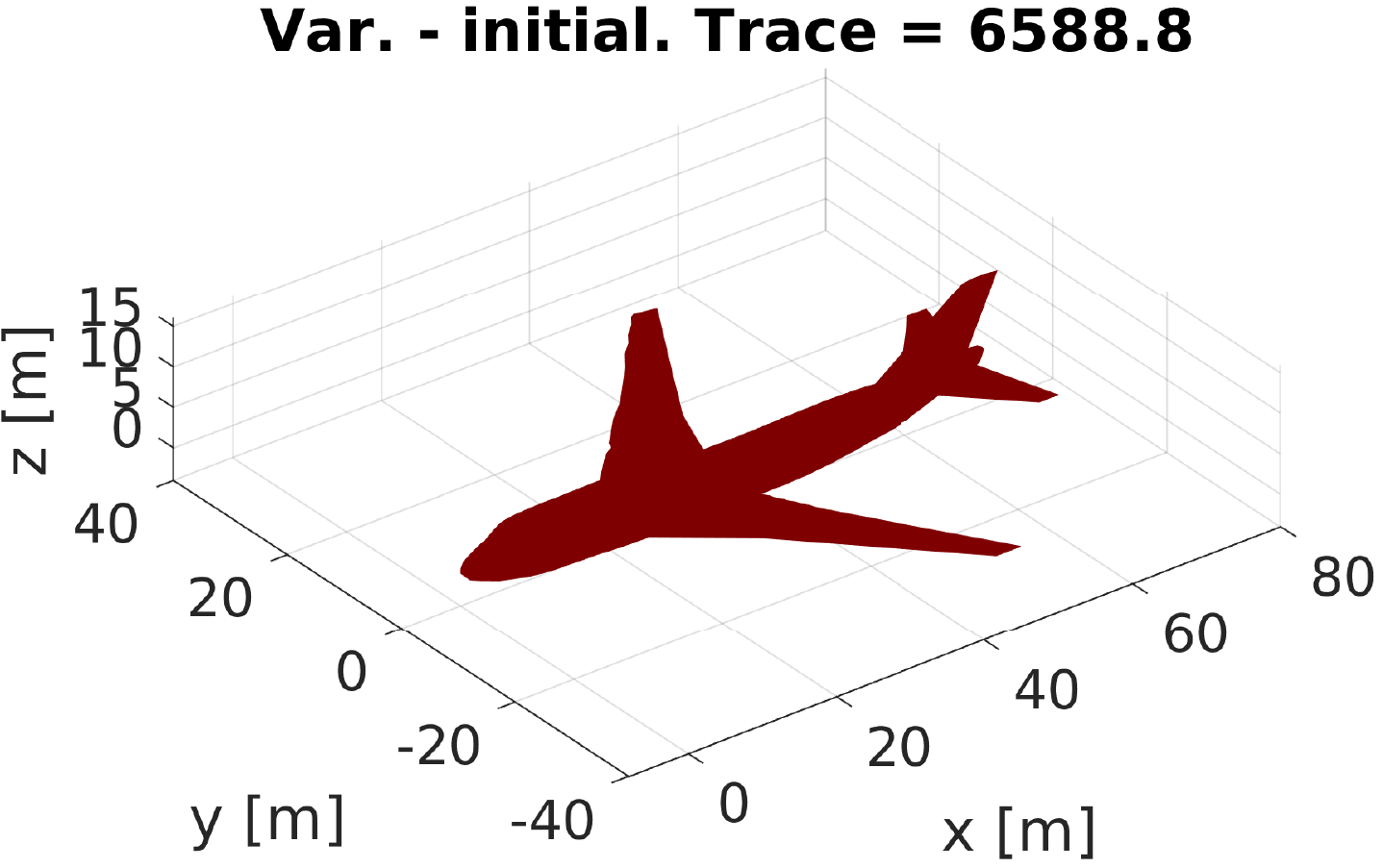}
	\caption{Map initial mean (top) and variance (bottom) at 0 s.}
	\label{subfig:boeing747_var_initial}
    \end{subfigure}
    \captionsetup[subfigure]{position=b}
    \begin{subfigure}{0.24\textwidth}
            \includegraphics[width=1.0\textwidth]{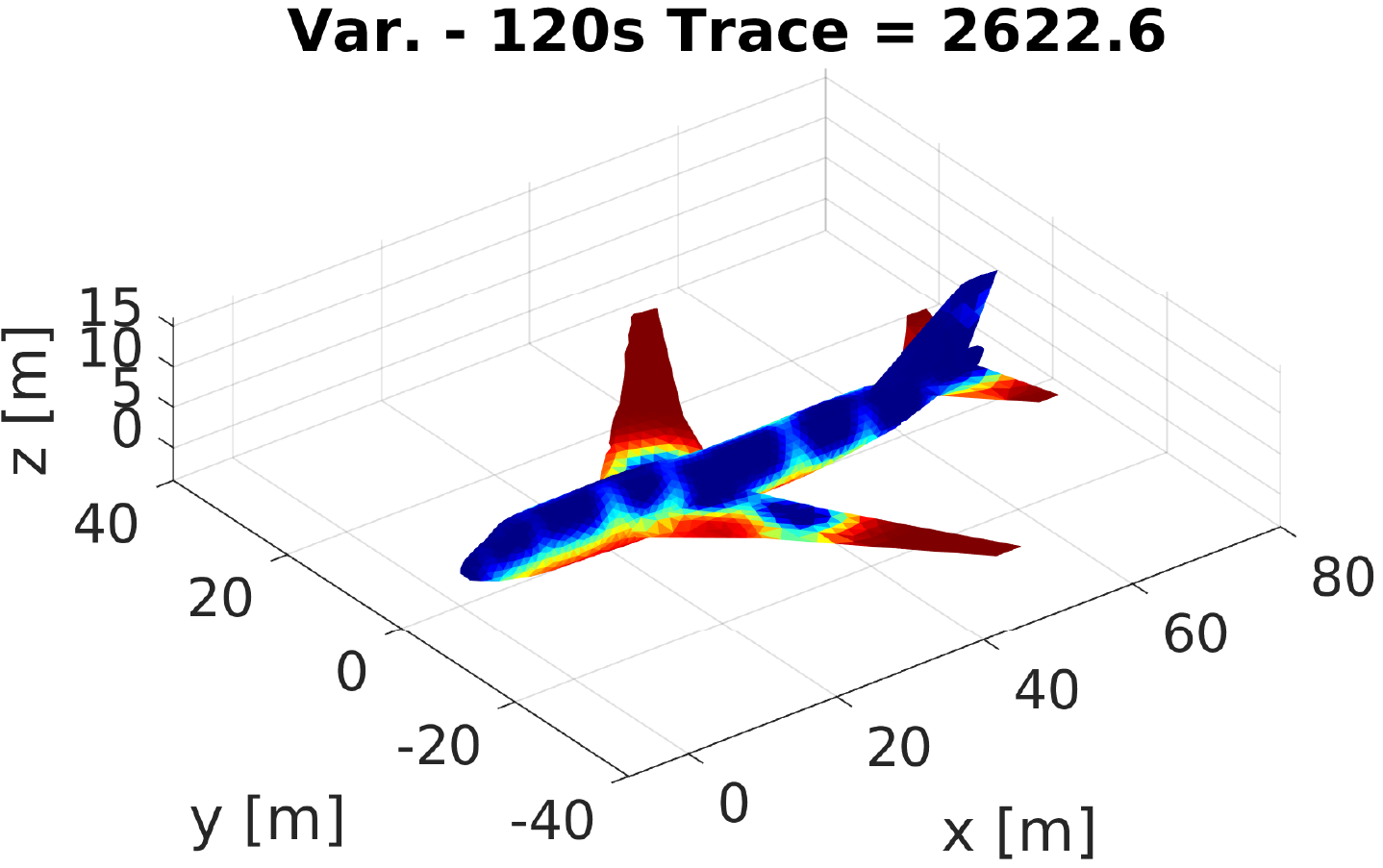}
	\caption{Map mean (top) and variance (bottom) at 120 s.}
	\label{subfig:boeing747_var_120}
    \end{subfigure}
    \captionsetup[subfigure]{position=b}
    \begin{subfigure}{0.24\textwidth}
            \includegraphics[width=1.0\textwidth]{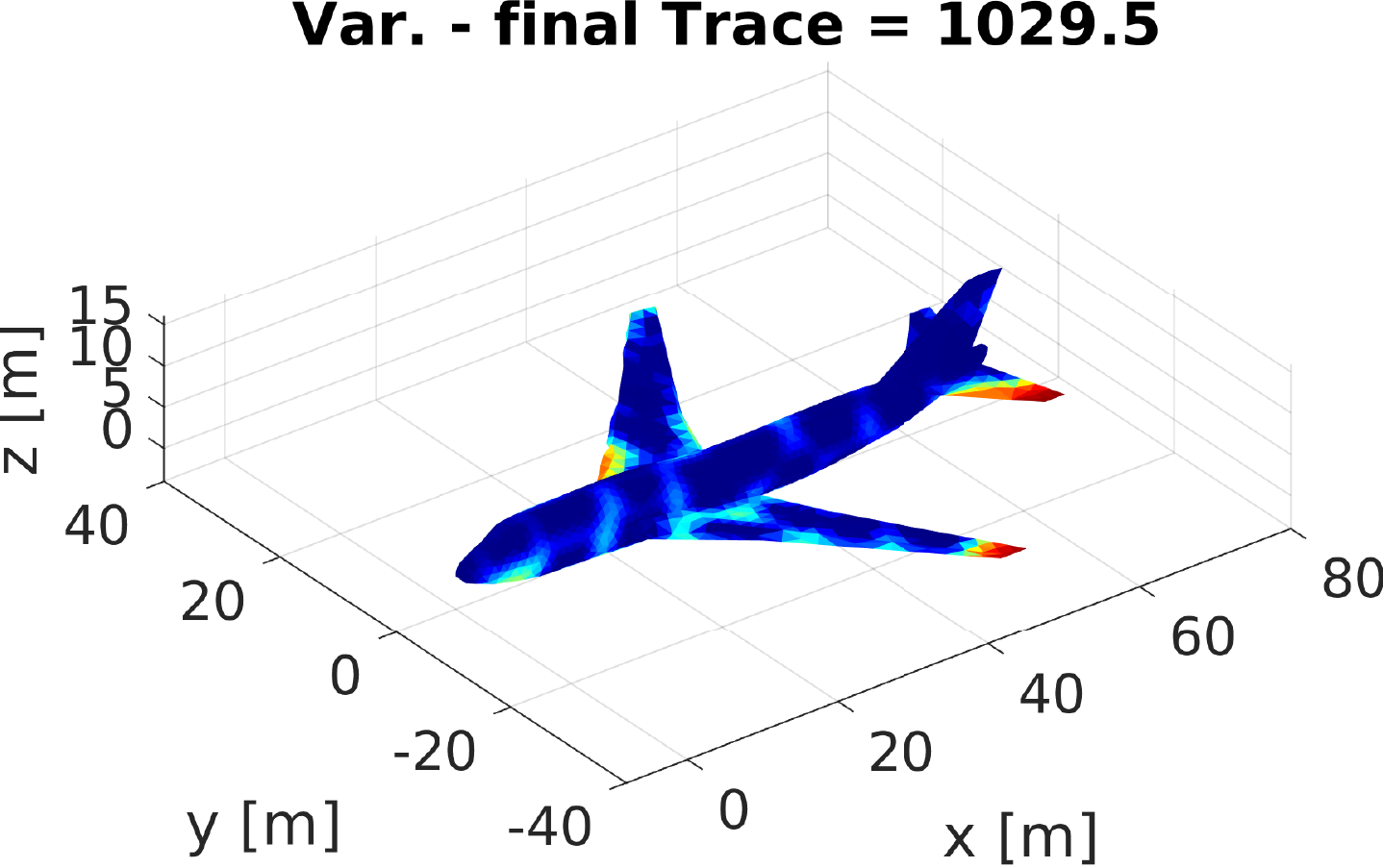}
	\caption{Map final mean (top) and variance (bottom) at 240 s.}
	\label{subfig:boeing747_var_final}
    \end{subfigure}
    \caption{Simulation results of the airplane inspection task using our proposed IPP method. }
    \label{fig:boeing747_inspection_process}%
\end{figure*}

\begin{figure}[t]
    \centering
    \captionsetup[subfigure]{position=b}
    \begin{subfigure}{0.23\textwidth}
            \includegraphics[width=1.0\textwidth]{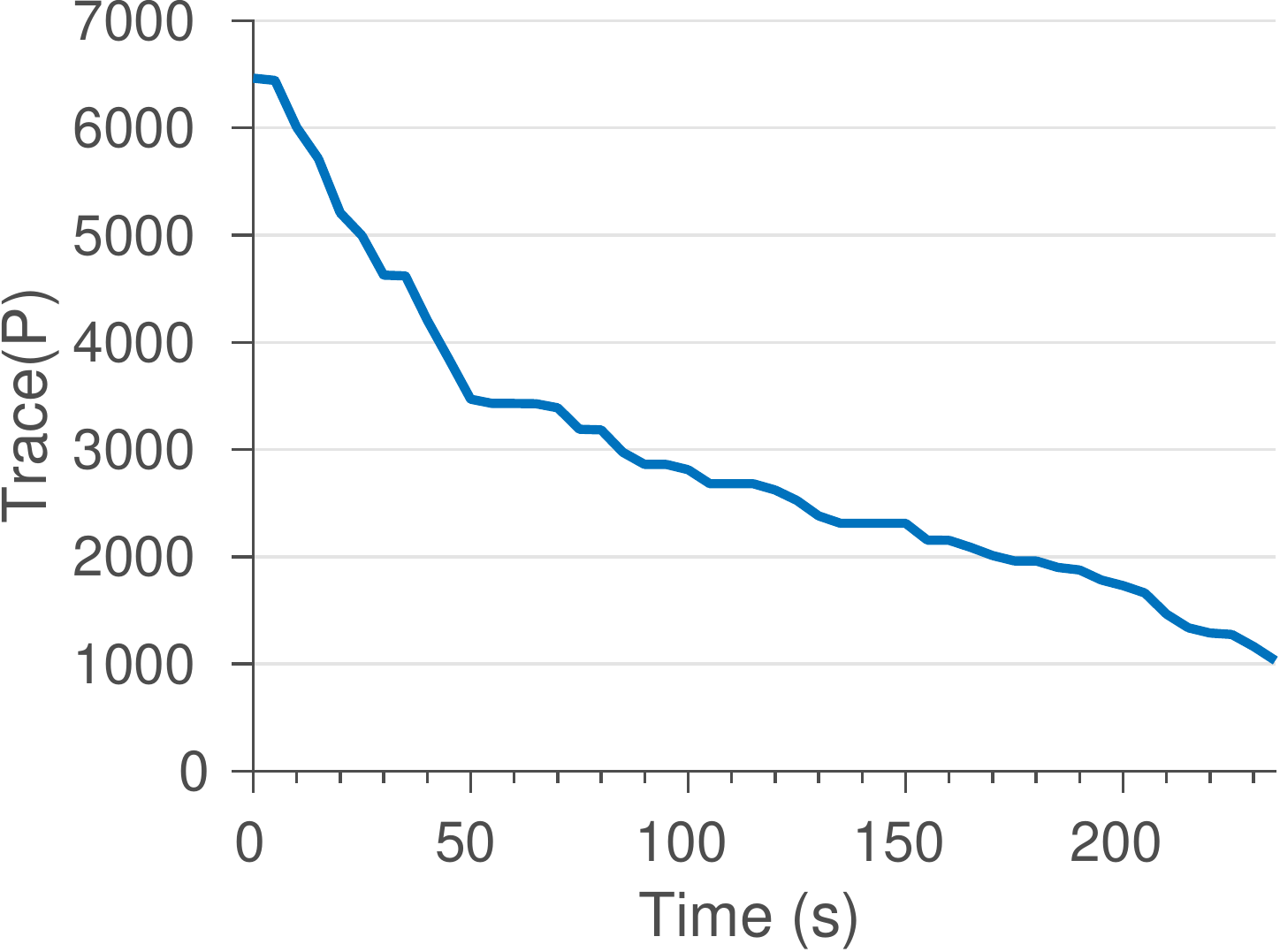}
	\caption{}
	\label{subfig:boeing747_P}
    \end{subfigure}
    \captionsetup[subfigure]{position=b}
    \begin{subfigure}{0.23\textwidth}
            \includegraphics[width=1.0\textwidth]{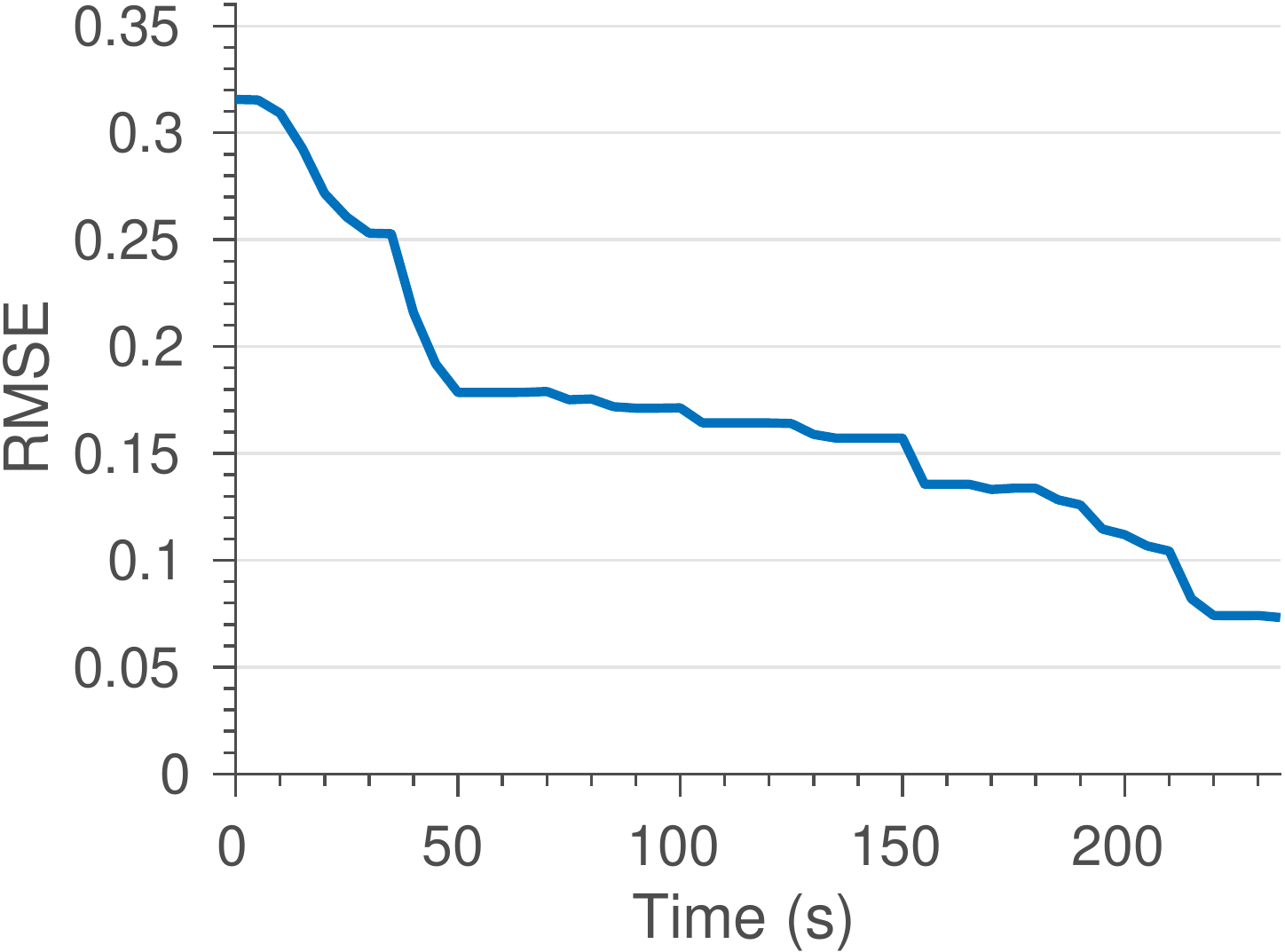}
	\caption{}
	\label{subfig:boeing747_RMSE}
    \end{subfigure}
    \caption{Map covariance uncertainty reduction and mapping error evolution of the airplane surface inspection task. 
    % (a) Trace of the map covariance. (b) Root mean squared error (RMSE) of the inspection results with respect to the ground truth.
    }
    \label{fig:boeing747_map_results}%
\end{figure}

\subsection{Spatial Correlations Evaluation}
We then evaluate the effects of taking spatial correlations of the information field into path planning. The same simulation scenario and setup as described in Section \ref{subsec:com_method} are used. We validate using mGPs to initialize the map covariance, which encodes spatial correlations. We also consider another two covariance initialization methods: a) an identity matrix, which indicates that the surface information on different regions are not correlated with each other; and b) a random semi-positive definite matrix (SPD) indicating the surface information has random spatial correlations. Then the same informative path planning and map update approaches are performed based on the three different map initialization methods. In implementation, the identity matrix and the random SPD are scaled to have the same covariance trace as the mGP method. For each method, 30 trials have been run. 

The evaluation results are shown in Fig. \ref{fig:correlation_eval}. It can be seen from the figure that by taking spatial correlations into path planning using our mGP mapping approach, the information gathering efficiency is significantly improved compared to not accounting for correlations (identity matrix) or assuming purely random correlations (random SPD). An interesting result is also observed in that the performance of considering no or purely random correlations are almost the same, indicating both mapping techniques lead to inefficient inspection of the surface. 

\subsection{Complex Surface Inspection}
Finally we present a more complex inspection case in which a UAV is required to map the surface temperature of a Boeing-747 airplane model. Similarly, we create a coarse 3D model of the airplane in SolidWorks and simulate a temperature distribution on its surface as the ground truth. A mesh of the surface is generated using Gmsh which has 1814 nodes and 3616 triangle facets. The total time budget for the mission is $B = 240$ s.

The simulation results are shown in Fig. \ref{fig:boeing747_inspection_process}, in which the inspection trajectory of the UAV and the surface map mean and variance during inspection are illustrated. As shown in Fig. \ref{subfig:boeing747_trajectory}, the UAV starts from the head of the airplane flying towards the tail to inspect the main body and then flies back to perform inspection around the two wings. It can be seen from Fig. \ref{subfig:boeing747_var_initial}-\ref{subfig:boeing747_var_final} that the UAV continuously reduces the surface map covariance uncertainty and mapping error, which is also illustrated in Fig. \ref{fig:boeing747_map_results}.

%% file: secs/07_conclusion.tex
\section{Conclusion and Future Work}\label{sec:conclsuion}

This paper presented an approach for active informative path planning (IPP) on 3D surfaces with a UAV. Our method applies a manifold Gaussian process with a geodesic kernel function to map surface information fields that captures spatial correlations of information over complex surface geometries. This model is integrated into a informative path planner that maximizes information gain within a planning budget while also respecting platform dynamic constraints and avoiding collisions. In a simulated temperature mapping scenario for a cylindrical storage tank, we showed that the proposed IPP approach could achieve faster map uncertainty and error reduction than the coverage path planner and a random inspection strategy. It was also shown that taking spatial correlations into planning using mGP for mapping could significantly improve the information gathering efficiency. Finally we validated the approach in a simulated airplane surface temperature inspection mission.  

Since the method is local, the online informative path planning may be trapped at some local minima. This might be resolved by combining the method with a global path planner. Future work shall also investigate inspecting dynamic changing information fields and multi-robot information gathering to further improve data acquisition efficiency.

%% file: main.bbl
\begin{thebibliography}{10}
\providecommand{\url}[1]{#1}
\csname url@rmstyle\endcsname
\providecommand{\newblock}{\relax}
\providecommand{\bibinfo}[2]{#2}
\providecommand\BIBentrySTDinterwordspacing{\spaceskip=0pt\relax}
\providecommand\BIBentryALTinterwordstretchfactor{4}
\providecommand\BIBentryALTinterwordspacing{\spaceskip=\fontdimen2\font plus
\BIBentryALTinterwordstretchfactor\fontdimen3\font minus
  \fontdimen4\font\relax}
\providecommand\BIBforeignlanguage[2]{{%
\expandafter\ifx\csname l@#1\endcsname\relax
\typeout{** WARNING: IEEEtran.bst: No hyphenation pattern has been}%
\typeout{** loaded for the language `#1'. Using the pattern for}%
\typeout{** the default language instead.}%
\else
\language=\csname l@#1\endcsname
\fi
#2}}

\bibitem{Gu2020}
W.~Gu, D.~Hu, L.~Cheng, Y.~Cao, A.~Rizzo, and K.~P. Valavanis, ``{Autonomous
  wind turbine inspection using a quadrotor},'' in \emph{2020 International
  Conference on Unmanned Aircraft Systems (ICUAS)}, 2020, pp. 709--715.

\bibitem{Chen2019}
S.~Chen, D.~F. Laefer, E.~Mangina, S.~M.~I. Zolanvari, and J.~Byrne, ``{UAV
  bridge inspection through evaluated 3d reconstructions},'' \emph{Journal of
  Bridge Engineering}, vol.~24, no.~4, p. 05019001, 2019.

\bibitem{Kim2015}
J.-W. Kim, S.-B. Kim, J.-C. Park, and J.-W. Nam, ``{Development of crack
  detection system with unmanned aerial vehicles and digital image
  processing},'' in \emph{Advances in Structure Engineering and Mechanics
  (ASEM15)}, 2015, pp. 25--29.

\bibitem{Miranda2019}
J.~Miranda, S.~Larnier, A.~Herbulot, and M.~Devy, ``{UAV-based inspection of
  airplane exterior screws with computer vision},'' in \emph{14th International
  Joint Conference on Computer Vision, Imaging and Computer Graphics Theory and
  Applications}, 2019, pp. 421--427.

\bibitem{Workswell2017}
Workswell, ``{Pipeline inspection with thermal diagnostics},'' Tech. Rep.,
  2017.

\bibitem{Bircher2016}
A.~Bircher, M.~Kamel, K.~Alexis, M.~Burri, P.~Oettershagen, S.~Omari,
  T.~Mantel, and R.~Siegwart, ``{Three-dimensional coverage path planning via
  viewpoint resampling and tour optimization for aerial robots},''
  \emph{Autonomous Robots}, vol.~40, no.~6, pp. 1059--1078, 2016.

\bibitem{bircher2017incremental}
A.~Bircher, K.~Alexis, U.~Schwesinger, S.~Omari, M.~Burri, and R.~Siegwart,
  ``An incremental sampling-based approach to inspection planning: the rapidly
  exploring random tree of trees,'' \emph{Robotica}, vol.~35, no.~6, pp.
  1327--1340, 2017.

\bibitem{Hitz2017}
G.~Hitz, E.~Galceran, M.-{\`{E}}. Garneau, F.~Pomerleau, and R.~Siegwart,
  ``{Adaptive continuous-space informative path planning for online
  environmental monitoring},'' \emph{Journal of Field Robotics}, vol.~34,
  no.~8, pp. 1427--1449, 2017.

\bibitem{Popovic2020}
M.~Popovi{\'{c}}, T.~Vidal-Calleja, G.~Hitz, J.~J. Chung, I.~Sa, R.~Siegwart,
  and J.~Nieto, ``{An informative path planning framework for UAV-based terrain
  monitoring},'' \emph{Autonomous Robots}, vol.~44, no.~6, pp. 889--911, 2020.

\bibitem{galceran2013survey}
E.~Galceran and M.~Carreras, ``A survey on coverage path planning for
  robotics,'' \emph{Robotics and Autonomous systems}, vol.~61, no.~12, pp.
  1258--1276, 2013.

\bibitem{Atkar2001}
P.~Atkar, H.~Choset, A.~Rizzi, and E.~Acar, ``{Exact cellular decomposition of
  closed orientable surfaces embedded in R3},'' in \emph{2001 IEEE
  International Conference on Robotics and Automation (ICRA)}, 2001, pp.
  699--704.

\bibitem{Acar2002}
E.~U. Acar, H.~Choset, A.~A. Rizzi, P.~N. Atkar, and D.~Hull, ``{Morse
  decompositions for coverage tasks},'' \emph{The International Journal of
  Robotics Research}, vol.~21, no.~4, pp. 331--344, 2002.

\bibitem{Hover2012}
F.~S. Hover, R.~M. Eustice, A.~Kim, B.~Englot, H.~Johannsson, M.~Kaess, and
  J.~J. Leonard, ``{Advanced perception, navigation and planning for autonomous
  in-water ship hull inspection},'' \emph{International Journal of Robotics
  Research}, vol.~31, no.~12, pp. 1445--1464, 2012.

\bibitem{Alexis2015}
K.~Alexis, C.~Papachristos, R.~Siegwart, and A.~Tzes, ``{Uniform coverage
  structural inspection path–planning for micro aerial vehicles},'' in
  \emph{2015 IEEE International Symposium on Intelligent Control (ISIC)}, 2015,
  pp. 59--64.

\bibitem{Englot2012}
B.~Englot and F.~S. Hover, ``{Sampling-based coverage path planning for
  inspection of complex structures},'' in \emph{Proceedings of the 22nd
  International Conference on Automated Planning and Scheduling (ICAPS)}, 2012,
  pp. 29--37.

\bibitem{Papadopoulos2013}
G.~Papadopoulos, H.~Kurniawati, and N.~M. Patrikalakis, ``{Asymptotically
  optimal inspection planning using systems with differential constraints},''
  in \emph{2013 IEEE International Conference on Robotics and Automation
  (ICRA)}, 2013, pp. 4126--4133.

\bibitem{Pito1999}
R.~Pito, ``{A solution to the next best view problem for automated surface
  acquisition},'' \emph{IEEE Transactions on Pattern Analysis and Machine
  Intelligence}, vol.~21, no.~10, pp. 1016--1030, 1999.

\bibitem{Connolly1985}
C.~Connolly, ``{The determination of next best views},'' in \emph{1985 IEEE
  International Conference on Robotics and Automation (ICRA)}, 1985, pp.
  432--435.

\bibitem{Bircher2018}
A.~Bircher, M.~Kamel, K.~Alexis, H.~Oleynikova, and R.~Siegwart, ``{Receding
  horizon path planning for 3D exploration and surface inspection},''
  \emph{Autonomous Robots}, vol.~42, no.~2, pp. 291--306, 2018.

\bibitem{Yoder2016}
L.~Yoder and S.~Scherer, ``{Autonomous exploration for infrastructure modeling
  with a micro aerial vehicle},'' in \emph{Springer Tracts in Advanced
  Robotics}, 2016, vol. 113, pp. 427--440.

\bibitem{williams2006gaussian}
C.~E. Rasmussen and C.~K.~I. Williams, \emph{Gaussian processes for machine
  learning}.\hskip 1em plus 0.5em minus 0.4em\relax MIT press Cambridge, MA,
  2006.

\bibitem{Meera2019}
A.~A. Meera, M.~Popovic, A.~Millane, and R.~Siegwart, ``{Obstacle-aware
  adaptive informative path planning for uav-based target search},'' in
  \emph{2019 International Conference on Robotics and Automation (ICRA)}, 2019,
  pp. 718--724.

\bibitem{Williams2007}
O.~Williams and A.~Fitzgibbon, ``Gaussian process implicit surfaces,'' 2006.

\bibitem{Kim2015GP}
S.~Kim and J.~Kim, ``{GPmap: A unified framework for robotic mapping based on
  sparse Gaussian processes},'' in \emph{Springer Tracts in Advanced Robotics},
  2015, vol. 105, pp. 319--332.

\bibitem{Lee2019}
B.~Lee, C.~Zhang, Z.~Huang, and D.~D. Lee, ``{Online continuous mapping using
  Gaussian process implicit surfaces},'' in \emph{2019 International Conference
  on Robotics and Automation (ICRA)}, 2019, pp. 6884--6890.

\bibitem{Wu2020}
L.~Wu, R.~Falque, V.~Perez-Puchalt, L.~Liu, N.~Pietroni, and T.~Vidal-Calleja,
  ``{Skeleton-based conditionally independent Gaussian process implicit
  surfaces for fusion in sparse to dense 3d reconstruction},'' \emph{IEEE
  Robotics and Automation Letters}, vol.~5, no.~2, pp. 1532--1539, 2020.

\bibitem{Dragiev2011}
S.~Dragiev, M.~Toussaint, and M.~Gienger, ``{Gaussian process implicit surfaces
  for shape estimation and grasping},'' in \emph{2011 IEEE International
  Conference on Robotics and Automation (ICRA)}, 2011, pp. 2845--2850.

\bibitem{Vidal-Calleja2014}
T.~Vidal-Calleja, D.~Su, F.~{De Bruijn}, and J.~V. Miro, ``{Learning spatial
  correlations for Bayesian fusion in pipe thickness mapping},'' in \emph{2014
  IEEE International Conference on Robotics and Automation (ICRA)}, 2014, pp.
  683--690.

\bibitem{Jayasumana2015}
S.~Jayasumana, R.~Hartley, M.~Salzmann, H.~Li, and M.~Harandi, ``{Kernel
  methods on Riemannian manifolds with Gaussian rbf kernels},'' \emph{IEEE
  Transactions on Pattern Analysis and Machine Intelligence}, vol.~37, no.~12,
  pp. 2464--2477, 2015.

\bibitem{Niu2019}
M.~Niu, P.~Cheung, L.~Lin, Z.~Dai, N.~Lawrence, and D.~Dunson, ``{Intrinsic
  Gaussian processes on complex constrained domains},'' \emph{Journal of the
  Royal Statistical Society: Series B (Statistical Methodology)}, vol.~81,
  no.~3, pp. 603--627, 2019.

\bibitem{Borovitskiy2020}
V.~Borovitskiy, A.~Terenin, P.~Mostowsky, and M.~P. Deisenroth, ``Matern
  gaussian processes on riemannian manifolds,'' in \emph{Advances in Neural
  Information Processing Systems}, 2020.

\bibitem{Sugiyama2008}
M.~Sugiyama, H.~Hachiya, C.~Towell, and S.~Vijayakumar, ``{Geodesic Gaussian
  kernels for value function approximation},'' \emph{Autonomous Robots},
  vol.~25, no.~3, pp. 287--304, 2008.

\bibitem{DelCastillo2015}
E.~del Castillo, B.~M. Colosimo, and S.~D. Tajbakhsh, ``{Geodesic Gaussian
  processes for the parametric reconstruction of a free-form surface},''
  \emph{Technometrics}, vol.~57, no.~1, pp. 87--99, 2015.

\bibitem{Sun2015}
L.~Sun, T.~Vidal-Calleja, and J.~V. Miro, ``{Bayesian fusion using
  conditionally independent submaps for high resolution 2.5D mapping},'' in
  \emph{2015 IEEE International Conference on Robotics and Automation (ICRA)},
  2015, pp. 3394--3400.

\bibitem{Kennedy1993}
H.~V. Kennedy, ``{Modeling noise in thermal imaging systems},'' in
  \emph{Infrared Imaging Systems: Design, Analysis, Modeling, and Testing IV},
  1993, pp. 66--77.

\bibitem{Mellinger2011}
D.~Mellinger and V.~Kumar, ``{Minimum snap trajectory generation and control
  for quadrotors},'' in \emph{2011 IEEE International Conference on Robotics
  and Automation (ICRA)}, 2011, pp. 2520--2525.

\bibitem{Richter2016}
C.~Richter, A.~Bry, and N.~Roy, ``{Polynomial trajectory planning for
  aggressive quadrotor flight in dense indoor environments},'' in
  \emph{Robotics Research}, 2016, vol. 114, pp. 649--666.

\bibitem{Adams2014}
W.~J. Adams and J.~H. Elder, ``{Effects of specular highlights on perceived
  surface convexity},'' \emph{PLoS Computational Biology}, vol.~10, no.~5, p.
  e1003576, 2014.

\bibitem{Geuzaine2009}
C.~Geuzaine and J.-F. Remacle, ``{Gmsh: A 3-D finite element mesh generator
  with built-in pre- and post-processing facilities},'' \emph{International
  Journal for Numerical Methods in Engineering}, vol.~79, no.~11, pp.
  1309--1331, 2009.

\bibitem{Hansen2007}
N.~Hansen, ``{The CMA evolution atrategy: A comparing review},'' in
  \emph{Towards a New Evolutionary Computation}.\hskip 1em plus 0.5em minus
  0.4em\relax Springer Berlin Heidelberg, 2007, vol. 102, no. 2006, pp.
  75--102.

\end{thebibliography}
